\documentclass[letterpaper, 10 pt, conference]{ieeeconf}  % Comment this line out if you need a4paper

\pdfoutput=1

\IEEEoverridecommandlockouts                              % This command is only needed if 
\usepackage{graphics} % for pdf, bitmapped graphics files
\usepackage{epsfig} % for postscript graphics files
\usepackage{mathptmx} % assumes new font selection scheme installed
\usepackage{times} % assumes new font selection scheme installed
\usepackage{amsmath} % assumes amsmath package installed
\usepackage{amssymb}  % assumes amsmath package installed

\usepackage{listings} %for listing source code
\usepackage{algpseudocode} %for listing pseudocode

\usepackage{wrapfig}
\usepackage{caption}
\usepackage{subcaption}

\usepackage{booktabs}
\usepackage[bookmarks=false]{hyperref}

\title{\LARGE \bf
From Monocular SLAM to Autonomous Drone Exploration
}

\author{Lukas von Stumberg$^{1}$, Vladyslav Usenko$^{1}$, Jakob Engel$^{2}$, J\"org St\"uckler$^{3}$, and Daniel Cremers$^{1}$% <-this % stops a space
\thanks{This work has been supported by ERC Consolidator Grant 3D Reloaded (649323).}% <-this % stops a space
\thanks{$^{1}$The authors are with the Computer Vision Group, Computer Science Institute 9,
        Technische Universit\"at M\"unchen, 85748 Garching, Germany
        {\tt\small \{stumberg, usenko, cremers\}@in.tum.de}}%
\thanks{$^{2}$The author is with Oculus Research, Redmond, USA
        {\tt\small jajuengel@gmail.com}}%
\thanks{$^{3}$The author is with the Computer Vision Group, Visual Computing Institute, RWTH Aachen University, 52074 Aachen, Germany
        {\tt\small stueckler@vision.rwth-aachen.de}}%
}

\begin{document}

%\newfloatcommand{capbtabbox}{table}[][\FBwidth]

\thispagestyle{empty} 
\onecolumn 
 
\begin{center} 
\noindent 
 
This paper has been accepted for publication in \emph{2017 European Conference on Mobile Robots}. 
 
\vspace{2em} 

DOI: \href{https://doi.org/10.1109/ECMR.2017.8098709}{10.1109/ECMR.2017.8098709} 
 
IEEE Xplore: \url{http://ieeexplore.ieee.org/document/8098709/} 
\end{center} 
\vspace{3em} 
 
\copyright2017 IEEE. Personal use of this material is permitted. Permission from IEEE must be obtained for all other uses, in any current or future media, including reprinting/republishing this material for advertising or promotional purposes, creating new collective works, for resale or redistribution to servers or lists, or reuse of any copyrighted component of this work in other works.
 
\twocolumn

\setcounter{page}{1}

\maketitle
\thispagestyle{empty}
\pagestyle{empty}

%%%%%%%%%%%%%%%%%%%%%%%%%%%%%%%%%%%%%%%%%%%%%%%%%%%%%%%%%%%%%%%%%%%%%%%%%%%%%%%%
\begin{abstract}

Micro aerial vehicles (MAVs) are strongly limited in their payload and power capacity.
In order to implement autonomous navigation, algorithms are therefore desirable that use sensory equipment that is as small, low-weight, and low-power consuming as possible.
In this paper, we propose a method for autonomous MAV navigation and exploration using a low-cost consumer-grade quadrocopter equipped with a monocular camera.
Our vision-based navigation system builds on LSD-SLAM which estimates the MAV trajectory and a semi-dense reconstruction of the environment in real-time.
Since LSD-SLAM only determines depth at high gradient pixels, texture-less areas are not directly observed so that previous exploration methods that assume dense map information cannot directly be applied.
We propose an obstacle mapping and exploration approach that takes the properties of our semi-dense monocular SLAM system into account.
In experiments, we demonstrate our vision-based autonomous navigation and exploration system with a Parrot Bebop MAV.

\end{abstract}

%%%%%%%%%%%%%%%%%%%%%%%%%%%%%%%%%%%%%%%%%%%%%%%%%%%%%%%%%%%%%%%%%%%%%%%%%%%%%%%%
\section{INTRODUCTION}

Most autonomous micro aerial vehicles (MAVs) to date rely on depth sensing through e.g. laser scanners, RGB-D or stereo cameras.
Payload and power capacity are, however, limiting factors for MAVs, such that sensing principles are desirable that require as little size, weight, and power-consumption as possible.

In recent work, we propose large-scale direct simultaneous localization and mapping (LSD-SLAM~\cite{engel14eccv}) with handheld monocular cameras in real-time.
This method tracks the motion of the camera towards reference keyframes and at the same time estimates semi-dense depth at high gradient pixels in the keyframe.
By this, it avoids strong regularity assumptions such as planarity in textureless areas.
In this paper, we demonstrate how this method can be used for obstacle-avoiding autonomous navigation and exploration for a consumer-grade MAV.
We integrate our approach on the recently introduced Parrot Bebop MAV, which comes with a 30\,fps high-resolution fisheye video camera and integrated attitude sensing and control.

\begin{figure}[t]
 \centering
 \includegraphics[width=0.99\linewidth]{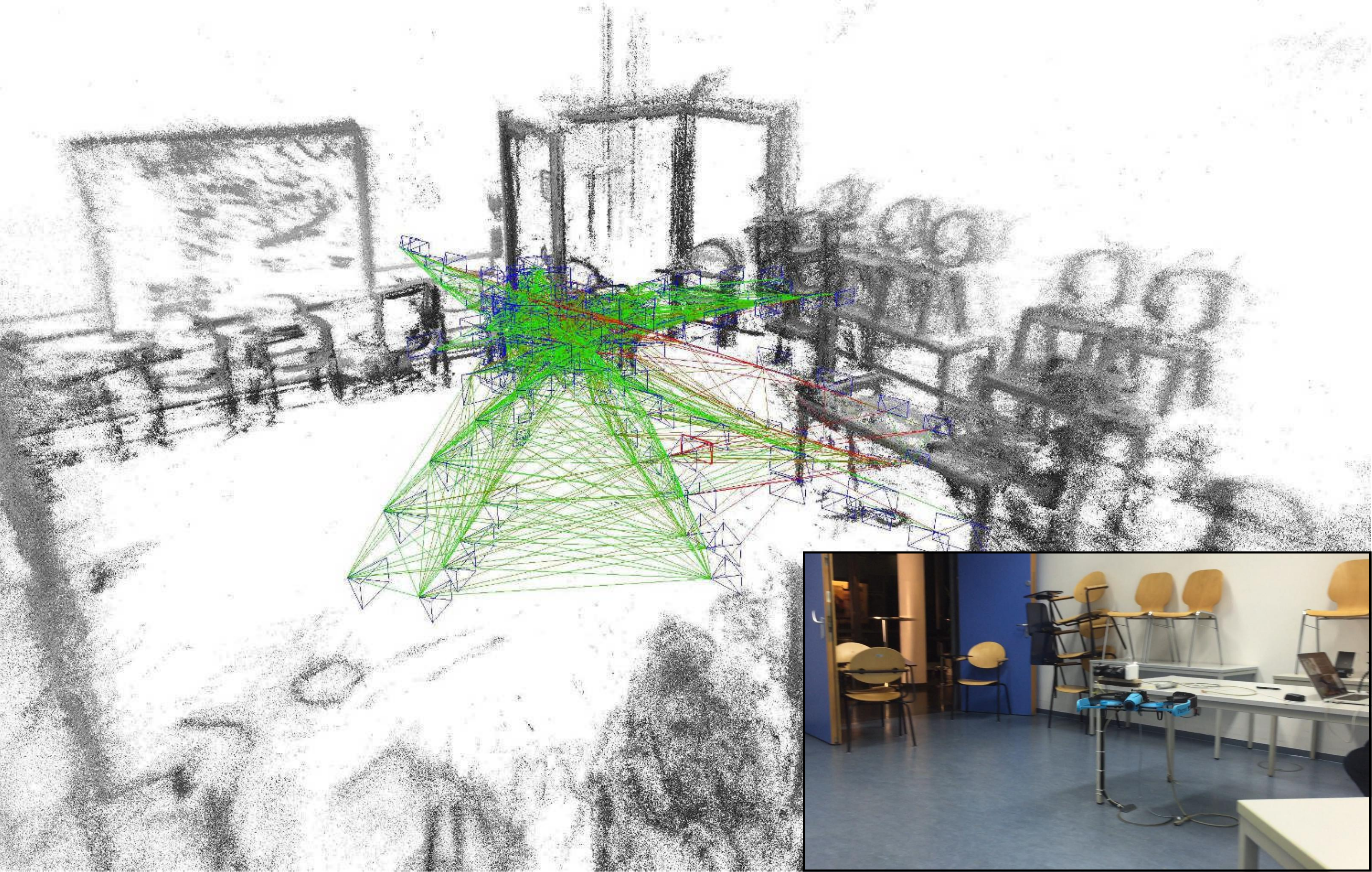}\\
 \caption{We propose vision-based autonomous navigation and exploration for small low-cost micro aerial vehicles equipped with monocular cameras. Our approach is based on large-scale direct SLAM which determines a semi-dense reconstruction of the environment.
 We integrate our approach on a commercially available Parrot Bebop MAV.} 
 \label{fig:teaser}
\end{figure}

Our proposed two-step exploration strategy is specifically and directly suited for semi-dense reconstructions as obtained with LSD-SLAM.
A simple but effective local exploration strategy, coined \emph{star discovery}, safely discovers free and occupied space in a local surrounding of a specific position in the environment.
In contrast to existing autonomous exploration approaches, our method takes the semi-dense depth measurement principle of LSD-SLAM based on motion parallax into account.
A global exploration strategy then determines interesting volume for further local explorations in order to sequentially discover novel parts of the environment.
We demonstrate the properties of our exploration strategy in several experiments with the Parrot Bebop.

\section{RELATED WORK}
Autonomous exploration by mobile robots has been investigated over many years, mainly relying on laser scanner sensors.
Yamauchi~\cite{yamauchi1997_frontier} proposed in his seminal work the so-called frontier-based exploration strategy that favors exploring the frontiers of the unexplored space in the map.
Some methods define a utility function~\cite{gonzalesbanos2002_utility_exploration,basilico2011_mcdm}, e.g., on paths or view poses that, for instance, trade-off discovered area with travel costs.
The approaches in~\cite{burgard2005_exploration,joho2007_3dexploration,stachniss05robotics} combine the probabilistic measure of information gain with travel cost in a measure of utility. 
Rekleitis et al.~\cite{rekleitis2012_slurm} optimize a utility function that favors the reduction of uncertainty in the map, and at the same time tries to achieve a fast exploration of the map.
All of the above methods rely on a dense map representation of the environment which is acquired using 2D and 3D laser range sensors.
In our case, these exploration methods are not directly applicable.
The exploration process needs to additionally consider how dense map information can be obtained from the visual semi-dense depth measurements of our SLAM system. 

Very recently, autonomous exploration by flying robots has attracted attention~\cite{shen2012_exploration,yoder2015_surface_exploration,nuske_jfr_2015,heng2015_mav_explore}.
Nuske et al.~\cite{nuske_jfr_2015} explore rivers using an MAV equipped with a continuously rotating 3D laser scanner.
They propose a multi-criteria exploration strategy to select goal points and traversal paths.
Heng et al.~\cite{heng2015_mav_explore} propose a two-step approach to visual exploration with MAVs using depth cameras.
Efficient exploration is achieved through maximizing information gain in a 3D occupancy map.
At the same time, high coverage of the viewed surfaces is determined along the path to the exploration pose. 
In order to avoid building up a dense 3D map of the environment and applying standard exploration methods, 
Shen et al.~\cite{shen2012_exploration} propose a particle-based frontier method that represents known and unknown space through samples.
This approach also relies on depth sensing through a 2D laser scanner and a depth camera.
Yoder and Scherer~\cite{yoder2015_surface_exploration} explore the frontiers of surfaces measured with a 3D laser scanner.
In~\cite{bircher2016_nbvplanner} a 3D occupancy map of the environment is acquired using an on-board depth sensor.
Next best views for exploration are selected by growing a random tree path planner in the free-space of the current map and choosing a branch to explore that maximizes the amount of unmapped space uncovered on the path.
Also these approaches use dense range or depth sensors which allow for adapting existing exploration methods from mobile robotics research.
Desaraju et al.~\cite{Desaraju2015} use a monocular camera and a dense motion stereo approach to find suitable landing sites of a UAV.

We propose an exploration method which is suitable for lightweight, low-cost monocular cameras.
Our visual navigation method is based on large-scale direct SLAM which recovers semi-dense reconstructions.
We take special care of the semi-dense information and its measurement process for obstacle mapping and exploration.

\section{AUTONOMOUS QUADROCOPTER NAVIGATION USING MONOCULAR LSD-SLAM}
\label{autonomous_nav}

We build on the TUM ARDrone package by Engel et al.~\cite{engel14ras} which has been originally developed for the Parrot ARDrone 2.0.
We transferred the software to the Parrot Bebop platform which comes with similar sensory equipment and onboard control. 
The Parrot Bebop is equipped with an IMU built from 3-axis magnetometer, gyroscope, and accelerometer.
It measures height using an ultrasonic sensor, an air pressure sensor and a vertical camera, similar to the Parrot ARDrone 2.0.
The MAV is equipped with a fisheye camera with wide 186$^\circ$ field-of-view.
The camera provides images at 30 frames per second. 
A horizontally stabilized region-of-interest is automatically extracted in software on the main processing unit of the MAV, and can be transmitted via wireless communication with the attitude measurements.

\subsubsection{State Estimation and Control}

The visual navigation system proposed in~\cite{engel14ras} integrates visual motion estimates from a monocular SLAM system with the attitude measurements from the MAV.
It filters both kinds of messages using a loosely-coupled Extended Kalman filtering (EKF) approach.
Since the attitude measurements and control commands are transmitted via wireless communication, they are affected by a time delay that needs to be compensated using the EKF framework.
Waypoint control of the MAV is achieved using PID control based on the EKF state estimate.
In monocular SLAM, the metric scale of motion and reconstruction estimates are not observable.
We probabilistically fuse ultrasonic and air pressure measurements and adapt the scale of the SLAM motion estimate to the observed metric scale~\cite{engel14ras}.

\subsubsection{Vision-Based Navigation Using Monocular LSD-SLAM}

LSD-SLAM~\cite{engel14eccv} is a keyframe based SLAM approach.
It maintains and optimizes the view poses of a subset of images, i.e. keyframes, extracted along the camera trajectory.
In order to estimate the camera trajectory, it tracks camera motion towards a reference keyframe through direct image alignment.
This requires depth in either of the images, which we estimate from stereo correspondences between the two images within the reference keyframe.
The poses of the keyframes are made globally consistent by mutual direct image alignment and pose graph optimization.

A key feature of LSD-SLAM is the ability to close trajectory loops within the keyframe graph.
In such an event, the view poses of the keyframes are readjusted to compensate for the drift that is accumulated through tracking along the loop.
This especially changes the pose of the current reference keyframe that is used for tracking, also inducing a change in the tracked motion estimate.
Yet, the tracked motion estimate is used to update the EKF that estimates the MAV state which is fed into the control loop.
At a loop closure, this visual motion estimate would update the filter with large erroneous velocities which would induce significant errors in the state estimate. 
In turn this could cause severe failures in flight control.
We therefore compensate for the changes induced by loop-closures with an additional pose offset on the visual estimate before feeding it into the EKF.

In order to initialize the system, the MAV performs a look-around maneuver in the beginning by flying a 360$^\circ$ turn on the spot while hovering up and down by several centimeters.
In this way, the MAV already obtains an initial keyframe map with a closed trajectory loop (Fig.~\ref{FigLsdBeforeAfter}).

\section{AUTONOMOUS OBSTACLE-FREE EXPLORATION WITH SEMI-DENSE DEPTH MAPS}

Autonomous exploration has been a research topic for many years targeting exploration of both 2D and 3D environments. 
In most 3D scenarios an exploration strategy works with a volumetric representation of the environment, such as a voxel grid or an octree, and uses laser-scanners or RGB-D cameras as sensors to build such a representation.

In this paper we devise an exploration strategy that builds on a fundamentally different type of sensor data -- 
semi-dense depth maps estimated with a single moving monocular camera. 
The difference to previously mentioned sensors lies in the fact that only for the image areas with strong gradients the depth can be estimated.
This means that especially initially during exploration, large portions of the map will remain unknown. 
The exploration strategy has to account for the motion parallax measurement principle of LSD-SLAM.

\subsection{Occupancy Mapping with Semi-Dense Depth Maps}\label{sec:occmapping}

In this work we use OctoMap~\cite{hornung2013_octomap} that provides an efficient implementation of hierarchical 3D occupancy mapping in octrees.
We directly use the semi-dense depth maps reconstructed with LSD-SLAM to create the 3D occupancy map. 
All keyframes are traversed and the measured depths are integrated via ray-casting using the camera model.

Since LSD-SLAM performs loop closures, the poses at which the depth maps of keyframes have been integrated into the map may change and the map will become outdated. 
We therefore periodically regenerate the map using the updated keyframe poses. 
While this operation may last for several seconds, the MAV hovers on the spot and waits until proceeding with the exploration.

Each voxel in the occupancy map stores the probability of being occupied in log-odds form.
In order to determine if a voxel is free or occupied, a threshold is applied on the occupancy probability (0.86 in our experiments).
During the integration of a depth measurement, all voxels along the ray in front of the measurement are updated with a probability value for missing voxels and measuring free-space.
The voxel at the depth measurement in turn is updated with a hit probability value.
Note that LSD-SLAM outputs not only the computed depths but also the variance of this estimate. Although measurements with  a high variance can be very noisy, they still contain information about the vicinity of the sensor. Therefore we insert only free space on a reduced distance for these pixels which assures that no wrong voxels are added.
Fig.~\ref{FigEx2Tree} shows an example occupancy map.

\begin{figure}[tb]
    \centering
    \begin{subfigure}[t]{0.46\linewidth}
        \includegraphics[height=0.12\textheight]{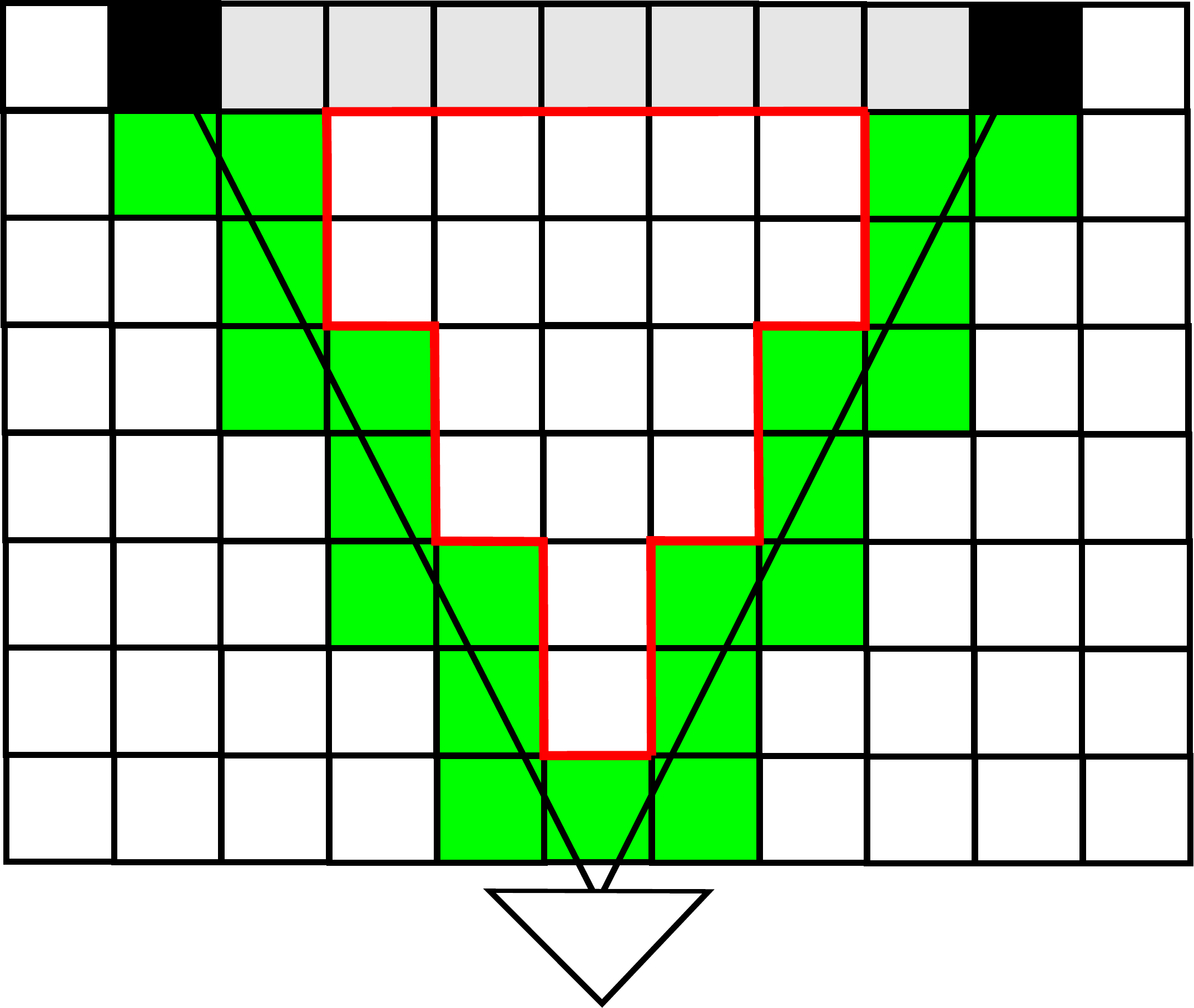}
        \label{FigSpikeBefore}
    \end{subfigure}
    \begin{subfigure}[t]{0.46\linewidth}
        \includegraphics[height=0.12\textheight]{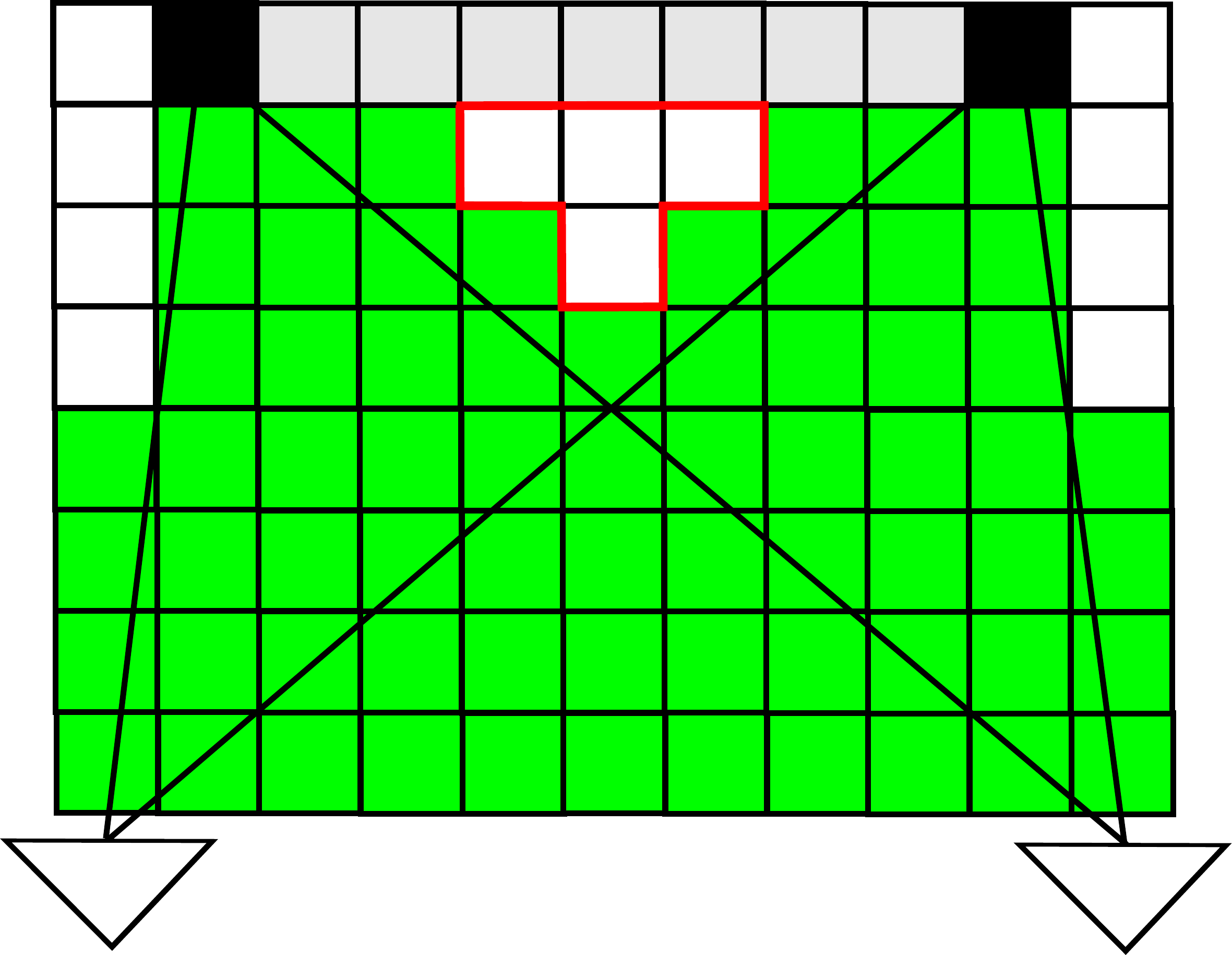}
        \label{FigSpikeAfter}
    \end{subfigure}
    \caption{With LSD-SLAM, only semi-dense depth measurements at edges and high-texture areas can be obtained. It provides no depth in textureless areas such as walls. This creates indentations of unknown space in the occupancy map (free space green, indentations red). Lateral motion (right picture) towards the measurable points, however, allows for reducing the indentations. }
	\label{FigSpikeEliminationStarDiscovery}
\end{figure}

\subsection{Optimal Motion for Exploration and Mapping}

\begin{figure}[tb]
    \centering
    \includegraphics[width=0.4\linewidth]{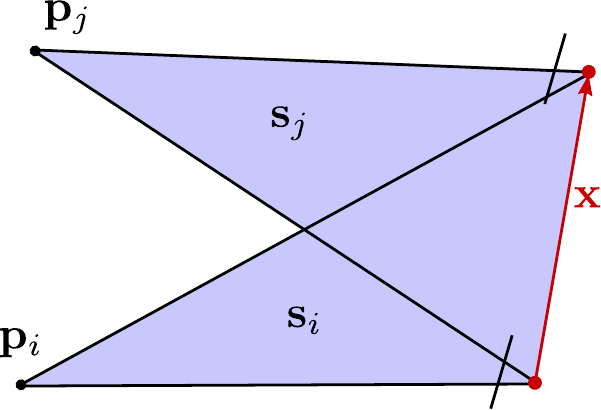}
    \caption{Discovered area~$\mathbf{s}_i$, $\mathbf{s}_j$ for measured points~$\mathbf{p}_i$, $\mathbf{p}_j$ in relation to the direction of motion~$\mathbf{x}$.}
	\label{optimal_direction}
\end{figure}

By using semi-dense reconstructions, we do not make strong assumptions such as planarity on the properties of the environment in textureless areas.
On the other hand, the use of semi-dense reconstruction in visual navigation leads to indentations of unknown volume which occur between the rays of free-space measured towards depth readings (Fig.~\ref{FigSpikeEliminationStarDiscovery}).
As we will analyze next, these indentations can be removed through lateral motion towards the measurable structures -- an important property that we will exploit in our exploration strategy.

Figure~\ref{optimal_direction} illustrates the problem of finding the direction of motion~$\mathbf{x} \in \mathbb{R}^3$ with~$\left\|\mathbf{x}\right\|_2 = 1$ such that it maximizes the observed free space in a 2D setting (without loss of generality). 
Assuming the camera center at the origin of the coordinate frame, the volume that is observed free in front of the measured point is cut by the triangle formed by the motion~$\mathbf{x}$ of the camera and the measured point~$\mathbf{p}_i$.
The magnitude of the vector
\begin{align}
    \mathbf{s}_i = \frac{1}{2} (\mathbf{p}_i \times \mathbf{x}) = \frac{1}{2} \hat{\mathbf{p}}_i \mathbf{x}
\end{align}
equals the observed free area,
where~$\hat{\mathbf{p}}_i$ is a skew-symmetric matrix formed from~$\mathbf{p}_i$ such that its product with~$\mathbf{x}$ corresponds to the cross-product between the two vectors.

To find the optimal direction we maximize (two times) the sum of squared areas of the triangles formed by all observed points,
\begin{align}
    S(\mathbf{x}) &= 2 \sum_{i=1}^{n} \mathbf{s}_i^\top \mathbf{s}_i = \frac{1}{2} \sum_{i=1}^{n} \mathbf{x}^\top \hat{\mathbf{p}}_i^\top \hat{\mathbf{p}}_i \mathbf{x},\\
    \left\|\mathbf{x}\right\|_2^2 &= \mathbf{x}^\top \mathbf{x} = 1.
\end{align}
Since we want to determine the optimal motion direction independent from its magnitude, we optimize the direction subject to a normalization constraint.

This constrained optimization problem can be solved using Lagrange multipliers,
\begin{align}
    L(\mathbf{x}, \lambda) &= S(\mathbf{x}) - \lambda (\mathbf{x}^\top \mathbf{x} - 1),
\end{align}
so that the optimal solution for~$\mathbf{x}$ should satisfy the equations
\begin{align}
    \nabla_\mathbf{x} L^\top &= \left( \sum_{i=1}^{n} \hat{\mathbf{p}}_i^\top \hat{\mathbf{p}}_i \right) \mathbf{x} - 2 \lambda \mathbf{x} = \mathbf{M} \mathbf{x} - 2 \lambda \mathbf{x} = \mathbf{0}, \\
    \frac{\partial L}{\partial \lambda} &= \mathbf{x}^\top \mathbf{x} - 1 = 0.
\end{align}
This implies that~$\mathbf{x}$ should be a (unit) eigenvector of the matrix~$\mathbf{M}$.
Moreover, the vector that corresponds to the largest eigenvalue produces the largest observation of the free space.

We perform Monte-Carlo simulations to further analyze the optimal motion direction.
Without any prior knowledge about the environment structure, we assume a uniform distribution of depths in the pixels.
We sample 600 points~$\mathbf{p}_i$ according to the following distribution: $u \sim U(0, 640), v \sim U(0,480), d \sim U(0.5, 5)$, where $(u,v)$ are the image coordinates, $d$ is the distance, and $U$ denotes uniform distribution. 
The points are reprojected into 3D space using the camera model and used for computing $\mathbf{M}$. 
Statistics for the eigenvalues and eigenvectors for 100 random simulations is accumulated in Table~\ref{tab:eigenvectors}. 
It demonstrates that the optimal direction for increasing the observed free space is a motion parallel to the image plane, i.e. sidewards or up-down motion.

\begin{table}
\caption{Summary of eigenvectors and eigenvalues of $\mathbf{M}$ over 100 random simulations where we draw image coordinates and depths from uniform distributions $u \sim U(0, 640)$, $v \sim U(0,480)$, $d \sim U(0.5, 5)$, respectively.}
\centering
 \begin{tabular}{cc} 
 \toprule
  \textbf{eigenvalue} &  \textbf{eigenvector} \\
 \midrule
 $7440.5 \pm 275.6$ & $(0.012, 0.996, -0.001) \pm (0.088, 0.005, 0.014)$ \\
 $7093.7 \pm 260.0$ & $(0.996, 0.012, -0.002) \pm (0.005, 0.088, 0.018)$ \\
 $1230.1 \pm 66.1$  & $(0.002, 0.001, 0.997) \pm (0.018, 0.015, 0.001)$ \\
 \bottomrule
\end{tabular}
\label{tab:eigenvectors}
\end{table}

\subsection{Obstacle-Free Local Exploration through Star Discoveries}
\label{localexploration}

\begin{figure}[tb]
  \begin{center}
    \includegraphics[width=0.43\linewidth]{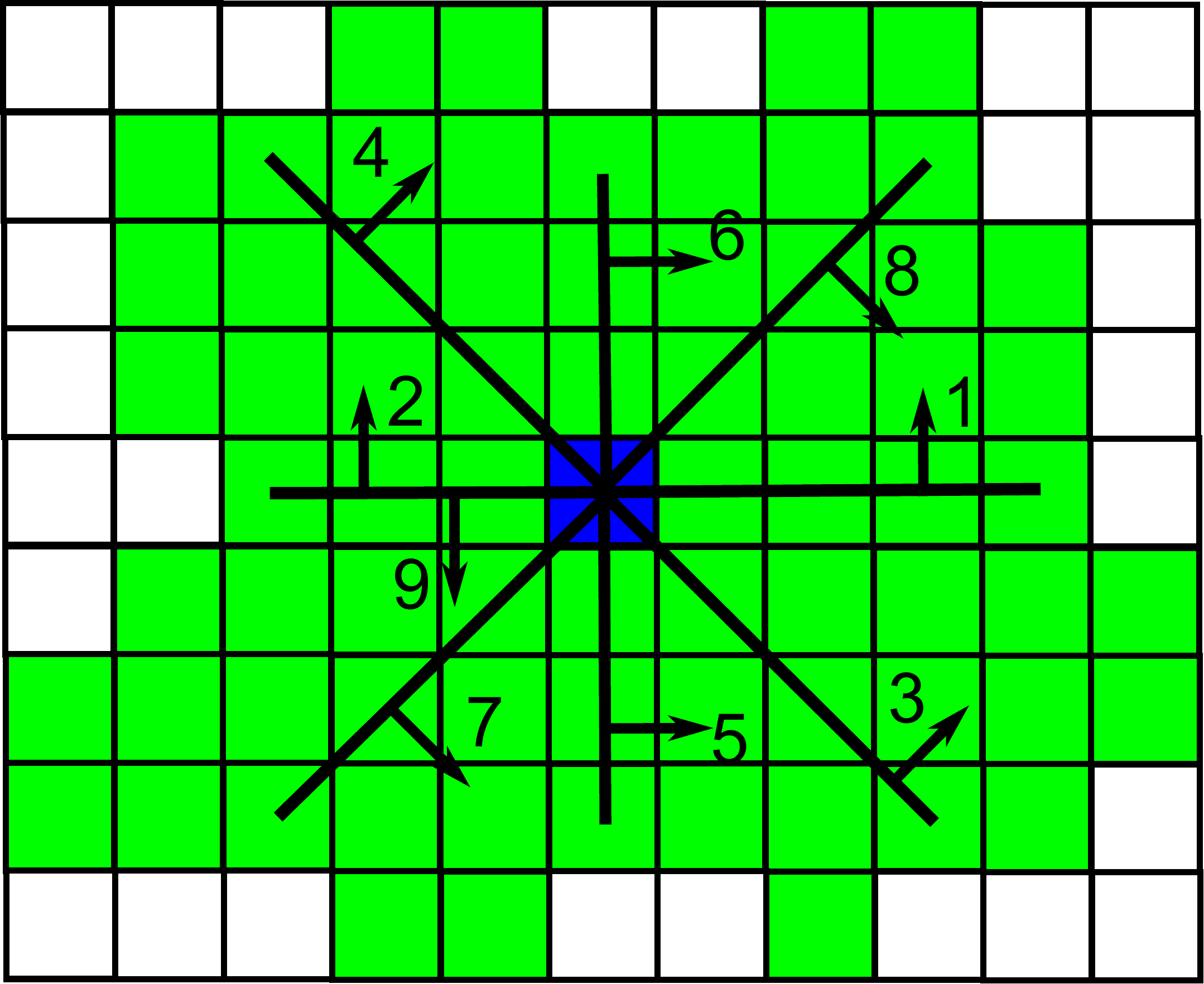}
  \end{center}
  \caption{Local star discovery exploration strategy. The MAV flies to the end of the lines in the order defined by the numbers, always facing the direction of the arrow corresponding to the line.}
\label{FigStarDiscovery}
\end{figure}

The characteristics of our semi-dense SLAM approach prevent the direct application of existing exploration approaches such as next-best view-planning or frontier-based exploration.
Frontiers of known space (occupied or measured free) occur at indentations of unknown space as well as between measured edges and textures on flat walls.
Simply flying to those boundaries would not allow discerning unknown from free or occupied space at the textureless boundaries as monocular SLAM requires motion parallax towards measurable structures and depth is only measured along the line-of-sight of semi-dense depth readings.
Next-best view planning aims at choosing a new view that maximizes the discovered unknown space at the new view pose.
Since measuring depth requires motion in our monocular SLAM system, it could be extended to measure the discovered space along the path to the new view point.
This procedure would be computationally very expensive, since for each potential view pose many ray-casting operations would need to be performed.
We propose a simpler but effective local exploration strategy that we call \emph{star discovery}, which discovers the indentations in the unknown volume around a specific position in the map.

In star discovery, the MAV flies a star-shape pattern (Fig.~\ref{FigStarDiscovery}).
In order to generate motion parallax for LSD-SLAM and to discover indentations in the unknown volume, the MAV flies with a 90$^\circ$ heading towards the motion direction.
Clearly, the MAV can only fly as far as the occupancy map already contains explored free-space.

The star-shape pattern is generated as follows:
We cast~$m$ rays from a specific position in the map at a predefined angular interval in order to determine the farest possible flight position along the ray.
The traversability of a voxel is determined by inflation of the occupied voxels by the size of the MAV.
In order to increase the success of the discovery, we perform this computation at $l$ different heights and choose the result of maximum size.

Only if the star discovery is fully executed, we redetermine the occupancy map from the updated LSD-SLAM keyframe map.
This also enables to postpone loop-closure updates towards the end of the exploration process, and provides a full 360$^\circ$ view from the center position of the star discovery.

Our exploration strategy is also favorable for the tracking performance of LSD-SLAM.
For instance, flying an outward facing ellipse of maximum size instead it could easily loose track because the MAV will only see few or no gradients when it flies close to an obstacle while facing it.

\subsection{Global Exploration}

\begin{figure}[tb]
    \centering
    \begin{subfigure}[t]{0.24\linewidth}
        \includegraphics[width=\textwidth]{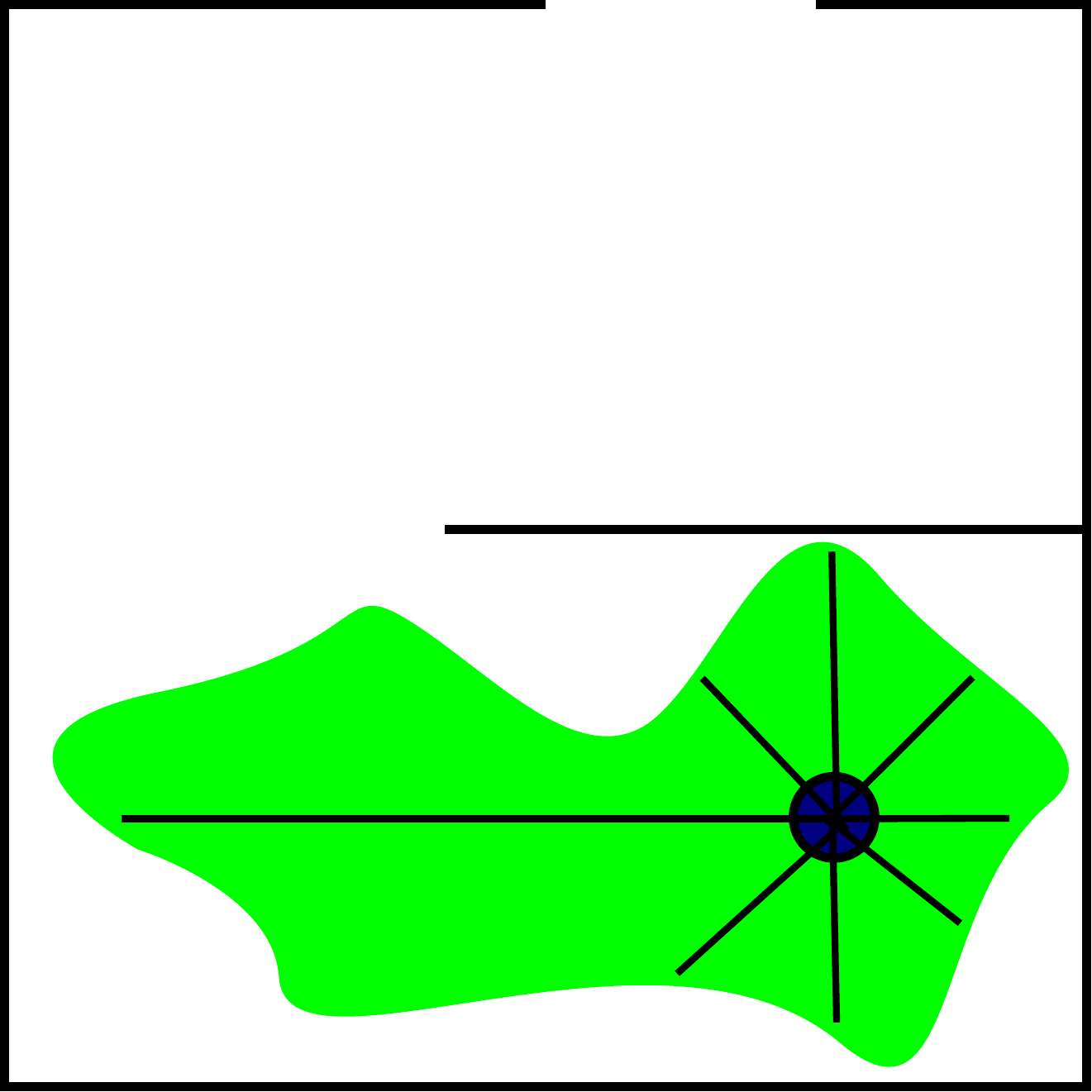}
        \caption{First}
%        \caption{The exploration starts at the blue circle. 
%        After an initial look-around, the green volume is marked as free.
%        The black lines illustrate the paths of the first star discovery. }
        \label{FigStep1}
    \end{subfigure}
    \begin{subfigure}[t]{0.24\linewidth}
        \includegraphics[width=\textwidth]{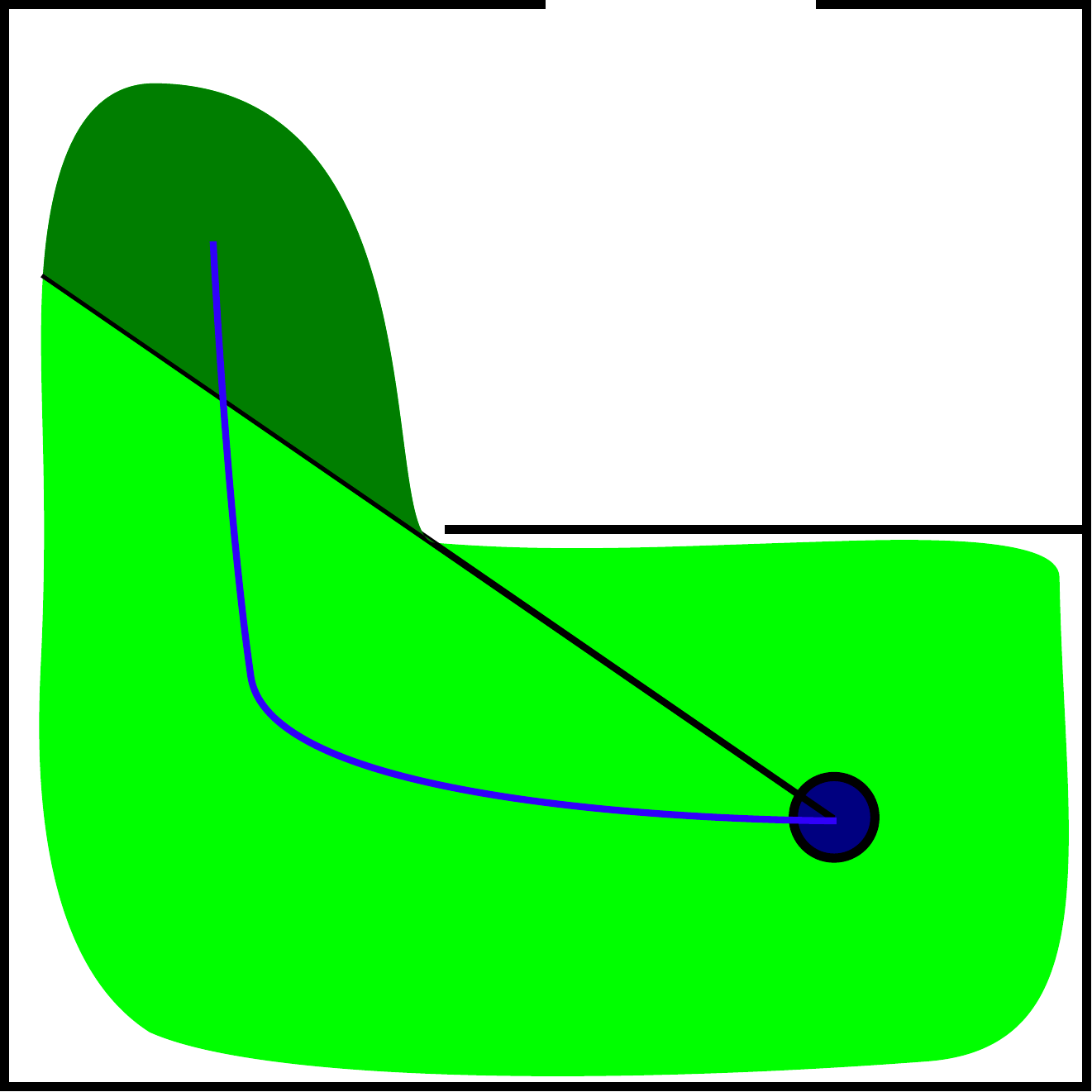}
        \caption{Second}
 %       \caption{All voxels that are free and in line-of-sight from the origin of the star discovery are marked light green.
%        The remaining voxels (dark green) are marked interesting. 
%        A path towards an interesting voxel is determined (blue line).}
        \label{FigStep2}
    \end{subfigure}
\begin{subfigure}[t]{0.24\linewidth}
        \includegraphics[width=\textwidth]{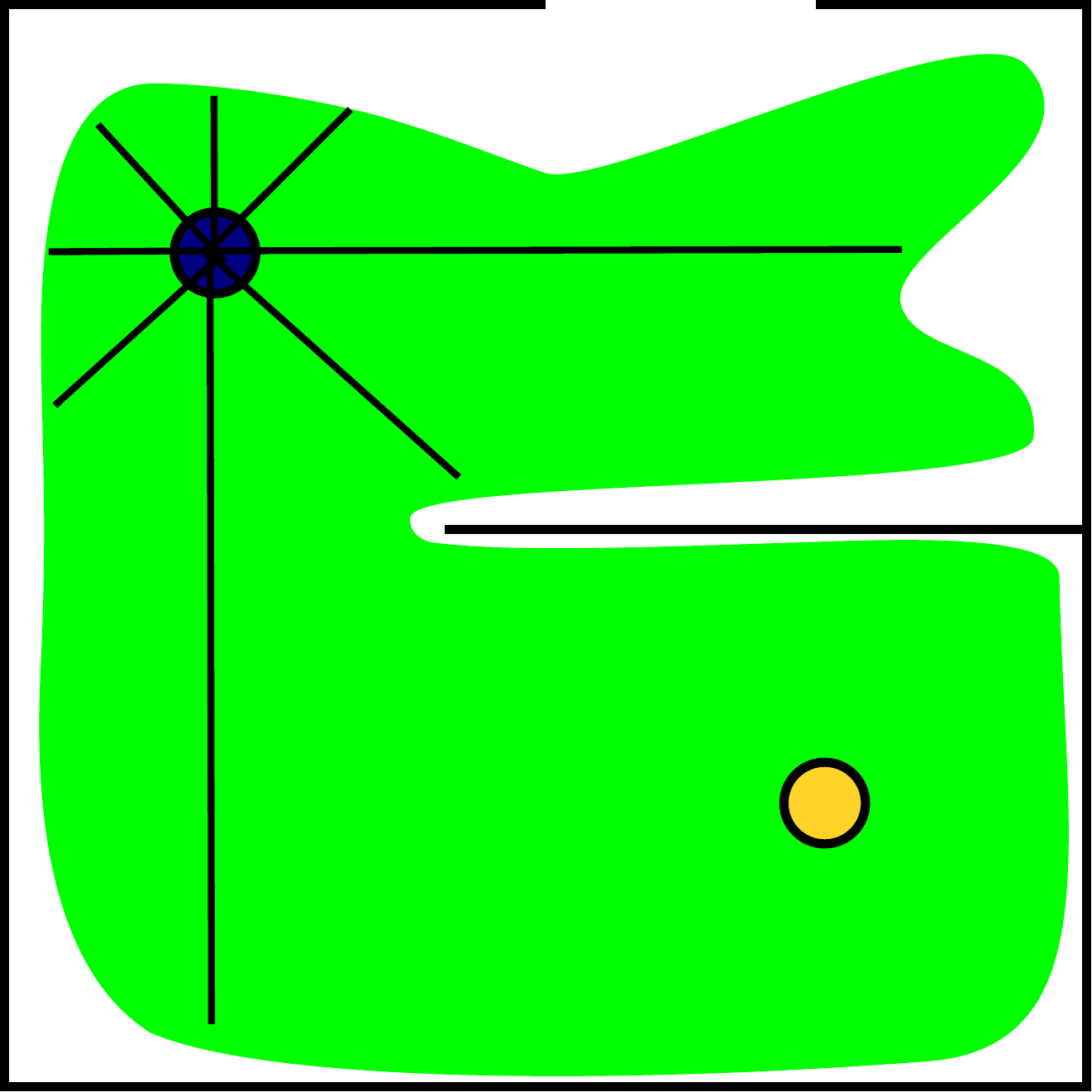}
        \caption{Third}
 %       \caption{A second star discovery (black lines) is executed at the new origin (blue).}
        \label{FigStep3}
    \end{subfigure}
    \begin{subfigure}[t]{0.24\linewidth}
        \includegraphics[width=\textwidth]{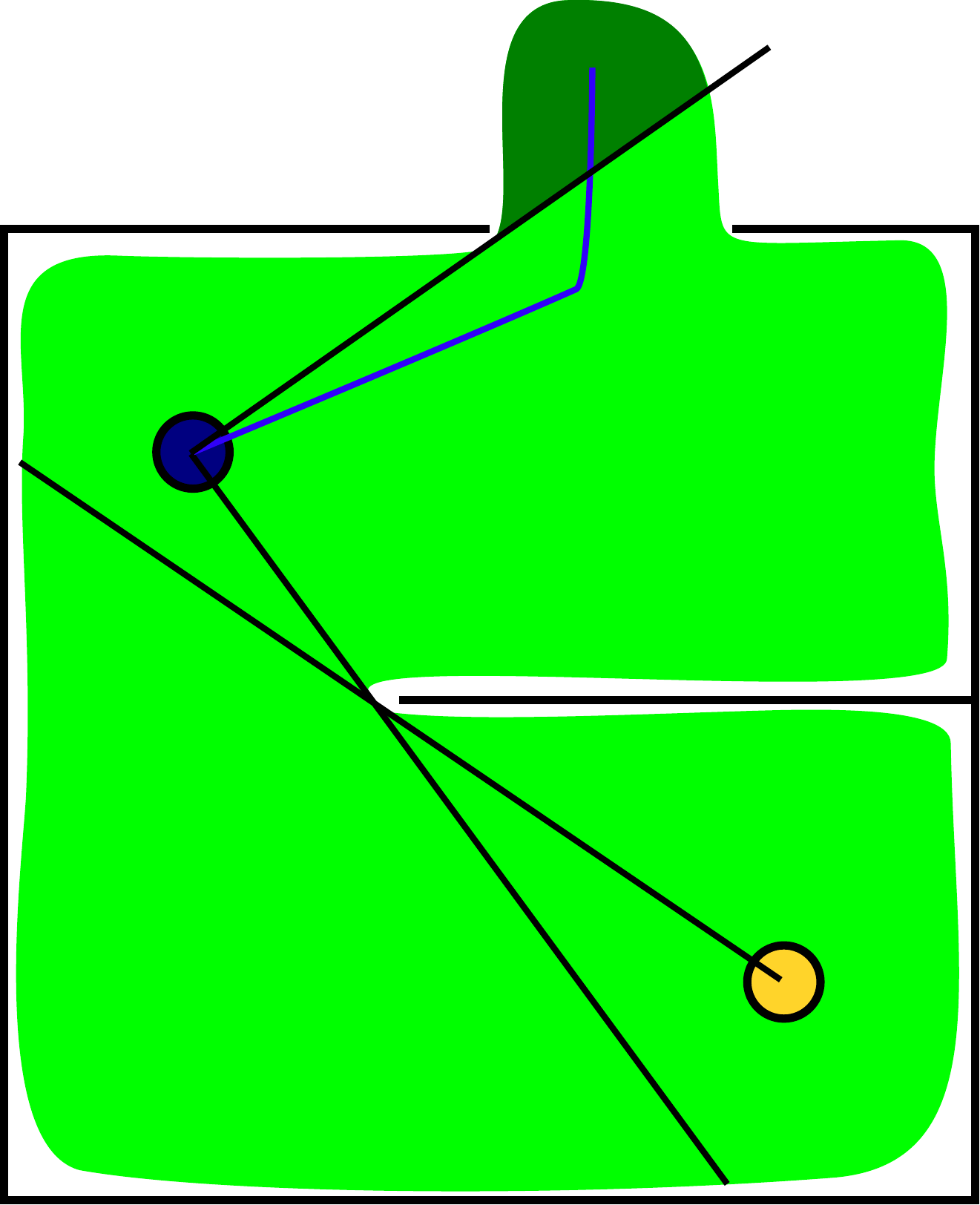}
        \caption{Fourth}
%        \caption{The dark green volume marks again the interesting volume. The algorithm finds a way out of the room.}
        \label{FigStep4}
    \end{subfigure}
    \caption{Four steps of the proposed exploration strategy. (a) The exploration starts at the blue circle. 
        After an initial look-around, the green volume is marked as free.
        The black lines illustrate the paths of the first star discovery. (b) All voxels that are free and in line-of-sight from the origin of the star discovery are marked light green.
        The remaining voxels (dark green) are marked interesting. 
        A path towards an interesting voxel is determined (blue line). (c) A second star discovery (black lines) is executed at the new origin (blue). (d) The dark green volume marks again the interesting volume. The algorithm finds a way out of the room. }
	\label{FigCompleteAlgorithm}
\end{figure}

Obviously, a single star discovery from one spot is not sufficient to explore arbitrarily shaped environments, as only positions on the direct line-of-sight from the origin can be reached (Fig.~\ref{FigCompleteAlgorithm}).
This induces a natural definition of interesting origins for subsequent star discoveries. 
We denote a voxel \emph{interesting} if it is known to be free but not in line-of-sight of any previous origin of star discovery. 

We determine the interesting voxels for starting a new star discovery as follows:
For every previously visited origin of a star discovery, we mark all free voxels in direct line-of-sight as visited. 
Then all free voxels in the inflated map are traversed and the ones that have not been marked are set to interesting.
With $m$ being the number of star discovery origins, the whole algorithm runs in $O(n^3 \cdot (m + hor^2 \cdot ver))$, where $hor$ and $ver$ are the number of voxels inflated in the horizontal and vertical directions.
We define $n$ as the number of voxels along the longest direction of the bounding box of the occupancy map.

Afterwards, we search a path in the occupancy map to one of the interesting voxels.
We look at several random voxels within the largest connected component of interesting voxels and choose the one from which we can execute the largest star discovery afterwards. 

As discussed above, frontier-based exploration would not be suitable with our monocular SLAM system, as the frontiers could be located on non-observable occupied structures such as textureless walls. 
In order to discover these structures, we propose star-shaped local exploration moves.
Our global exploration strategy determines new origins for these moves where freespace has been measured behind semi-dense structures that are not on the direct line-of-sight from the previous star discovery origin.

\section{RESULTS}

We evaluate our approach on a Parrot Bebop MAV in two differently sized and furnished rooms (a lab and a seminar room).
We recommend viewing the accompanying video of the experiments at \href{https://youtu.be/fWBsDwBJD-g}{https://youtu.be/fWBsDwBJD-g}.

\subsection{Experiment Setup}

We transmit the live image stream of the horizontally stabilized images of the Bebop to a control station PC via wireless communication.
The images are then processed on the control station PC to implement vision-based navigation and exploration based on LSD-SLAM.
All experiments were executed completely autonomous.

We report results of two experiments. 
The first experiment has been conducted in a lab room and demonstrates our star discovery exploration strategy.
In this simpler setting, we neglected depth measurements with high variance estimates and plan the star discovery only in a single height.
The second experiment evaluates local (star discovery) and global exploration strategies in a larger seminar room.
We enhanced the system  in several ways towards the first experiment to cope with the larger environment. 
We use depth measurements with high variance estimates as free space measurements for occupancy mapping as described in Sec.~\ref{sec:occmapping} instead of neglecting them.
In order to increase the possible coverage of the star discovery it is computed on several different heights and the one with the largest travel distance is used. 
When computing interesting voxels we as well use multiple heights for the center points. 
The robustness of the star discovery was improved by slightly reducing the maximum distance to the origin and by sending intermediate goals to the PID-controller. 
Finally, we start LSD-SLAM only right after takeoff, we improved the accuracy of the scale estimation and we readjusted the parameters of the PID-controller, the autopilot and the look-around maneuver.

\subsection{Qualitative Evaluation}

\subsubsection{Star Discovery}

\begin{figure}[tb]
    \centering
    \begin{subfigure}[b]{0.95\linewidth}
        \includegraphics[width=\linewidth]{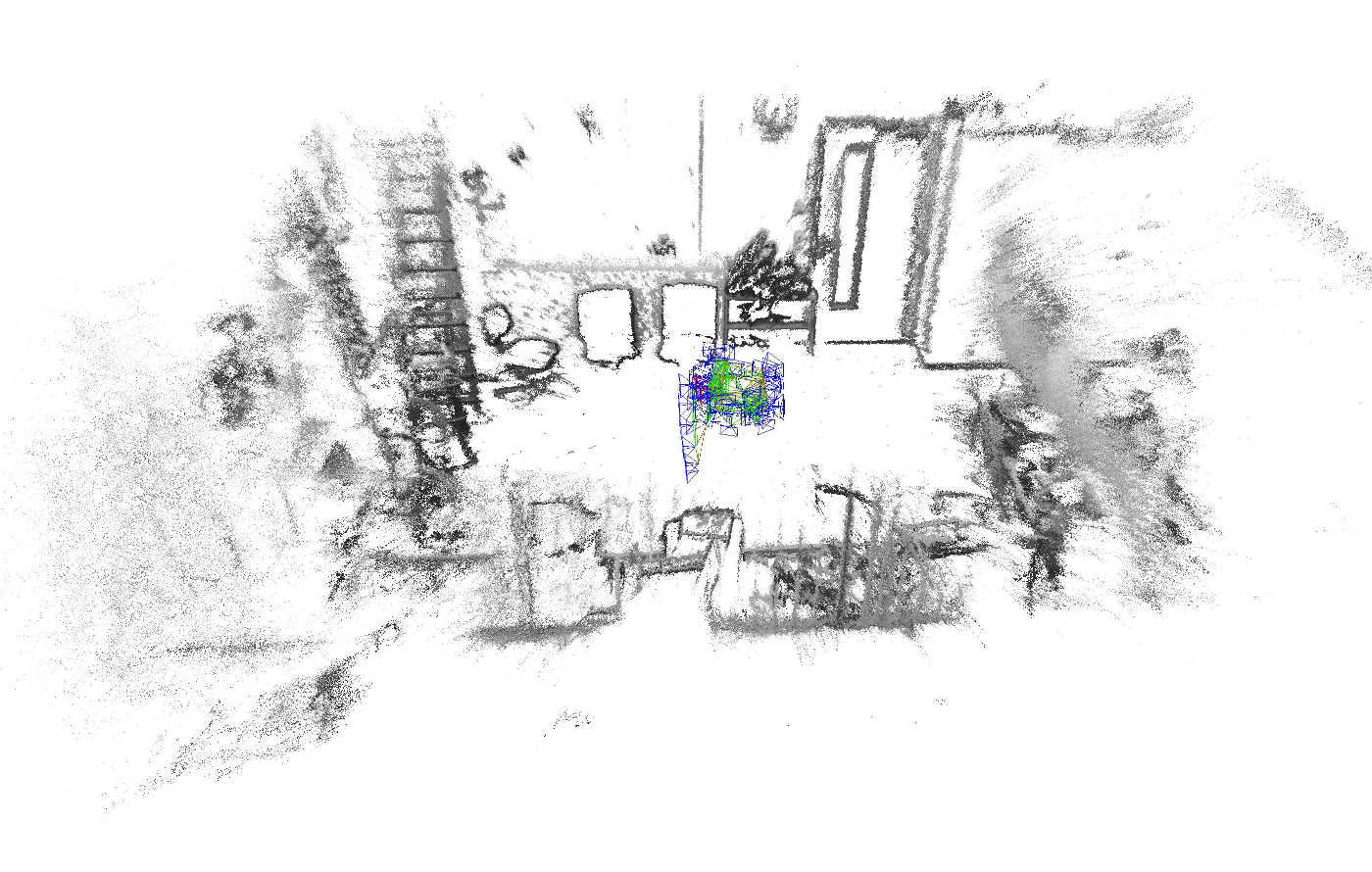}
    \end{subfigure}
    \caption{Semi-dense reconstruction after the look-around in the first experiment.}
	\label{FigLsdBeforeAfter} 
\end{figure}

\begin{figure}[tb]
    \centering
    ~
     \ \\
     \ \\

    \begin{subfigure}[b]{0.95\linewidth}
        \includegraphics[width=\linewidth]{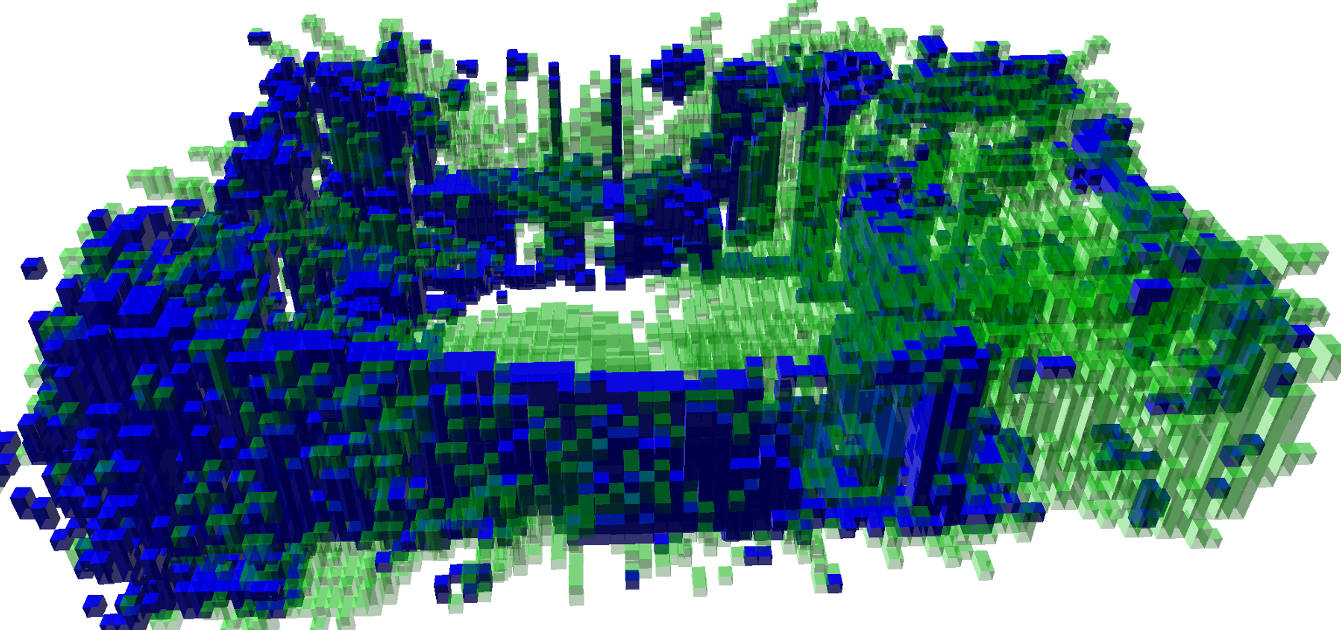}
        %\caption{free voxels that were unknown before the star discovery}
        
    \end{subfigure}

    ~ %add desired spacing between images, e. g. ~, \quad, \qquad, \hfill etc. 
    \caption{The difference between 3D occupancy map before and after star discovery. Occupied voxels are shown blue, free voxels that were unknown before the star discovery are green.}
	\label{FigOctomapBeforeAfter}
\end{figure}

\begin{figure}[tb]
    \centering
 \begin{subfigure}[b]{0.45\linewidth}
        \includegraphics[width=\linewidth]{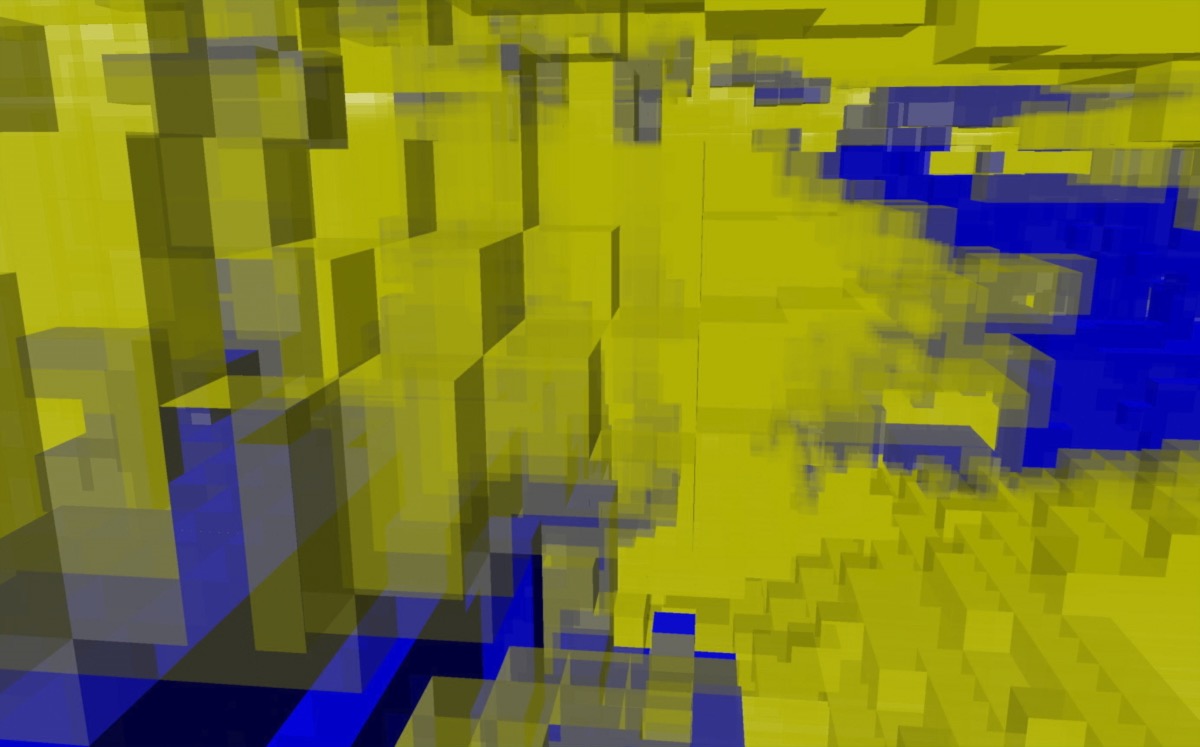}
        \caption{before star discovery}
        \label{FigSpike1Before}
    \end{subfigure}
    ~ %add desired spacing between images, e. g. ~, \quad, \qquad, \hfill etc. 
    \begin{subfigure}[b]{0.45\linewidth}
        \includegraphics[width=\linewidth]{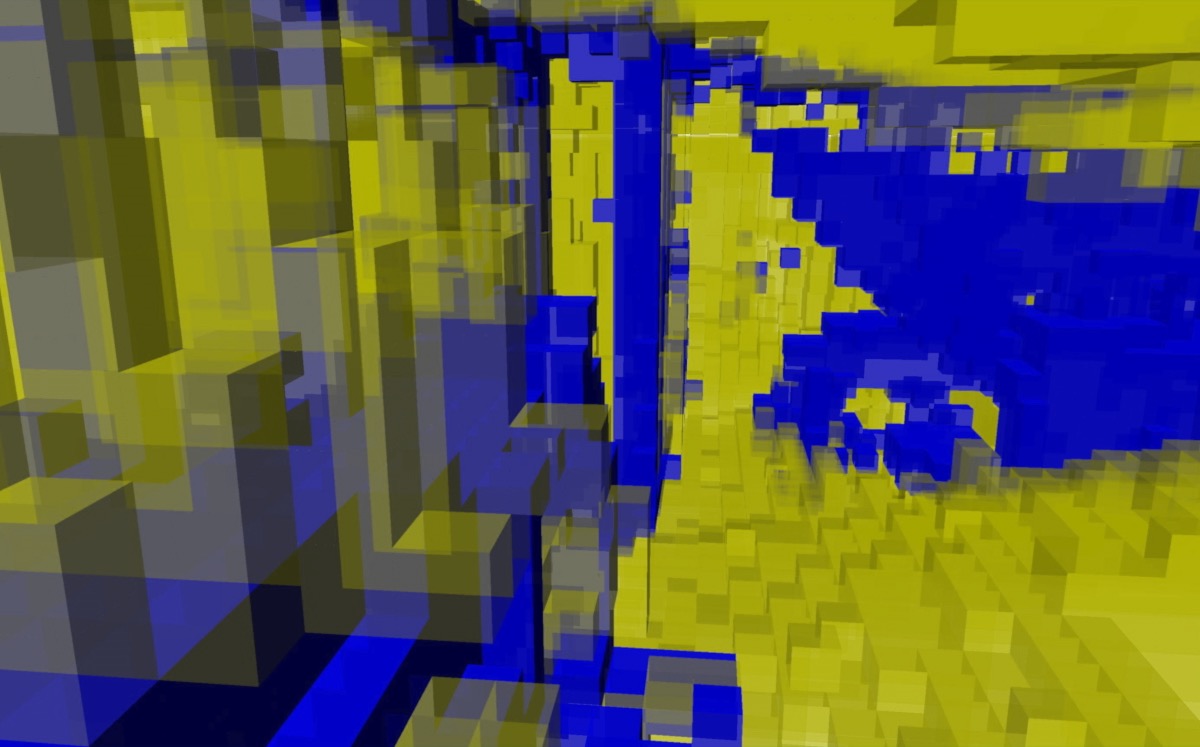}
        \caption{after star discovery}
        \label{FigSpike1After}
    \end{subfigure}
    \ \\
    \ \\
  \begin{subfigure}[b]{0.45\linewidth}
        \includegraphics[width=\linewidth]{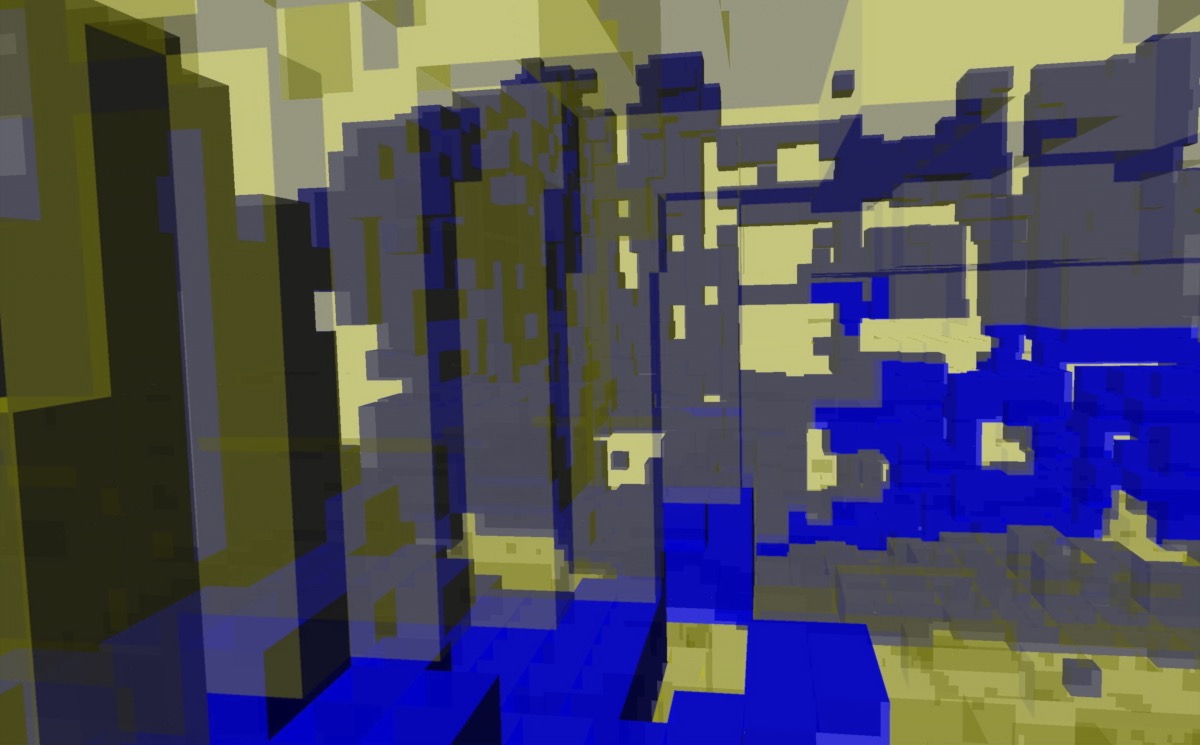}
        \caption{before star discovery}
        \label{FigSpike2Before}
    \end{subfigure}
    ~ %add desired spacing between images, e. g. ~, \quad, \qquad, \hfill etc. 
    \begin{subfigure}[b]{0.45\linewidth}
        \includegraphics[width=\linewidth]{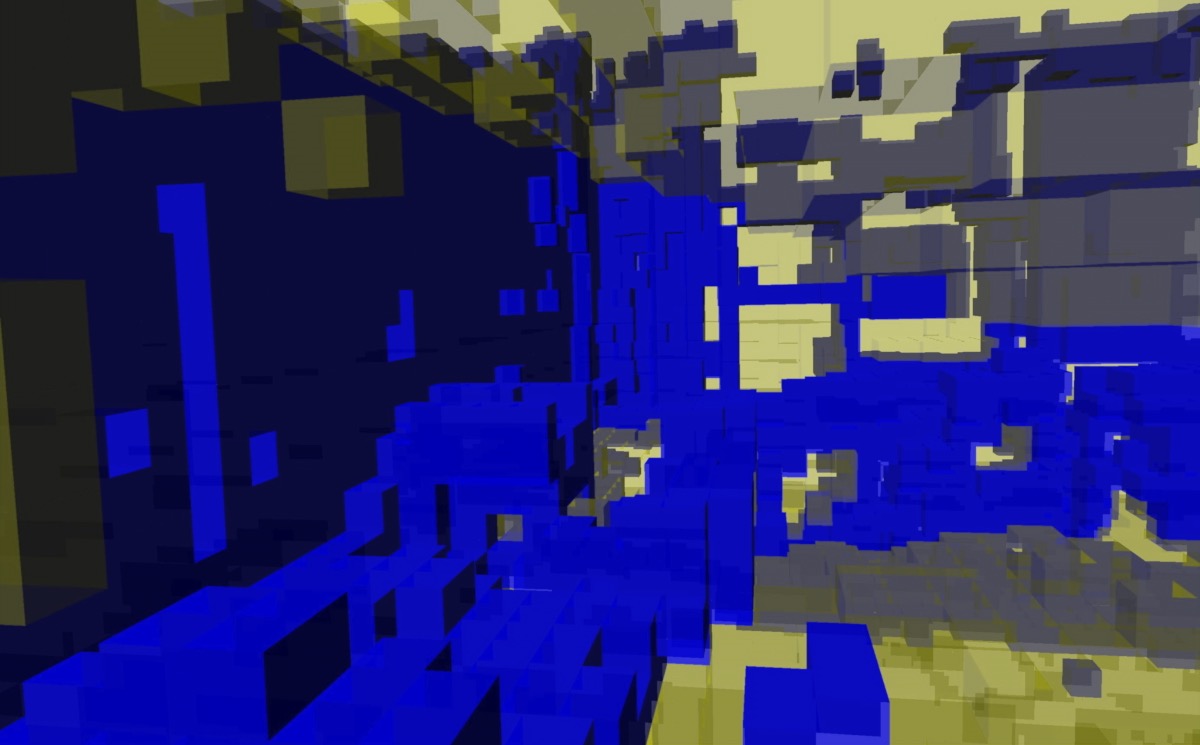}
        \caption{after star discovery}
        \label{FigSpike2After}
    \end{subfigure}
    ~ %add desired spacing between images, e. g. ~, \quad, \qquad, \hfill etc. 
    \caption{Indentations in unknown volume before and after the star discovery in the first experiment. Occupied voxels are blue and unknown voxels are yellow. The number of unknown (yellow) voxels is reduced through the discovery.}
	\label{FigSpikes1}
\end{figure}

\begin{figure}[tb]
 \begin{center}
    \includegraphics[width=0.99\linewidth]{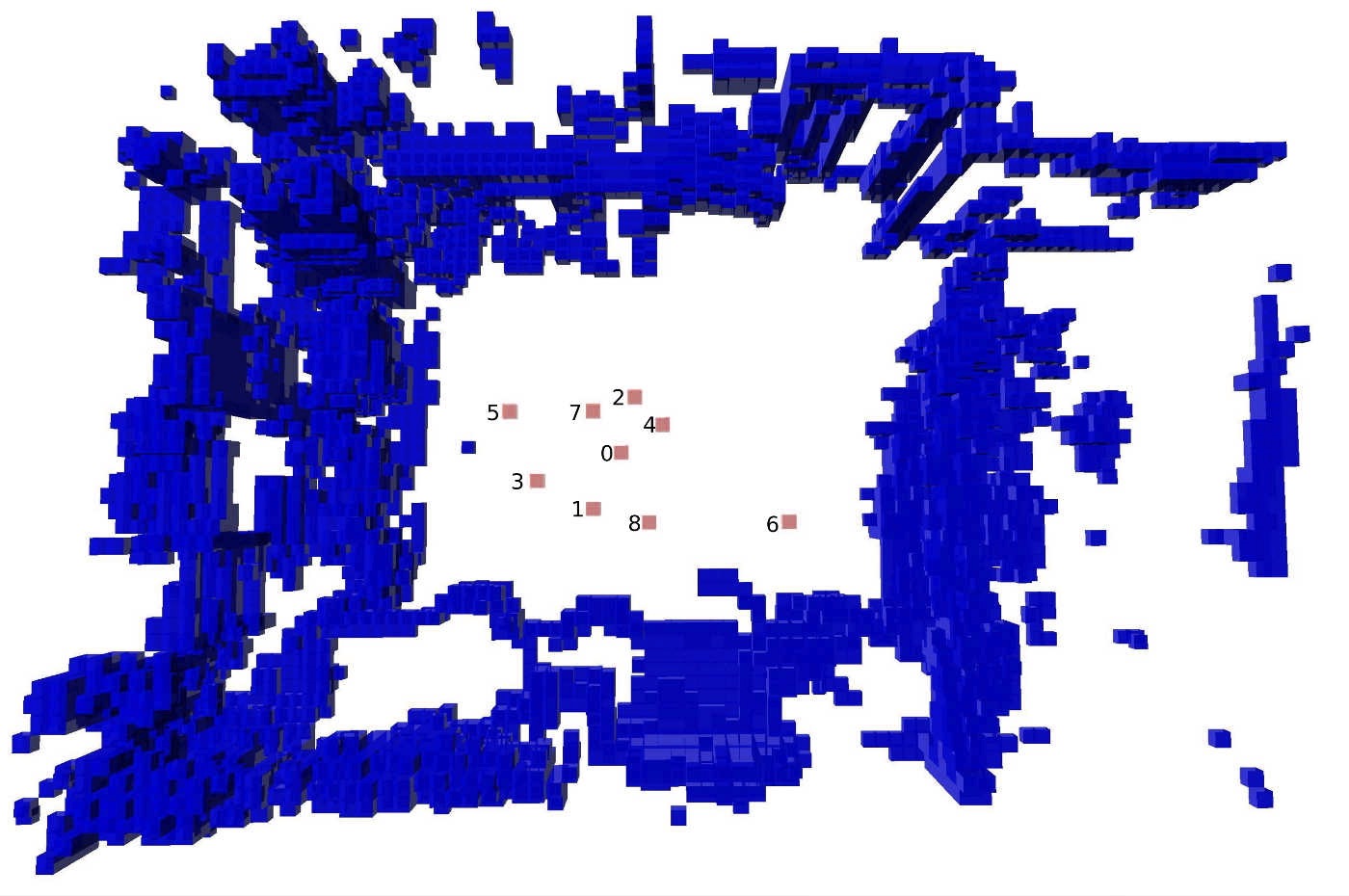}
 \end{center}
 \caption{Exploration plan of the star discovery in the first experiment. Occupied voxels are blue, approached voxels are red and numbered according to their approach order (0: origin).}
\label{FigDiscoveryPlan}
\end{figure}

\begin{figure}[tb]
    \centering
 \begin{subfigure}[b]{0.99\linewidth}
        \includegraphics[width=\linewidth]{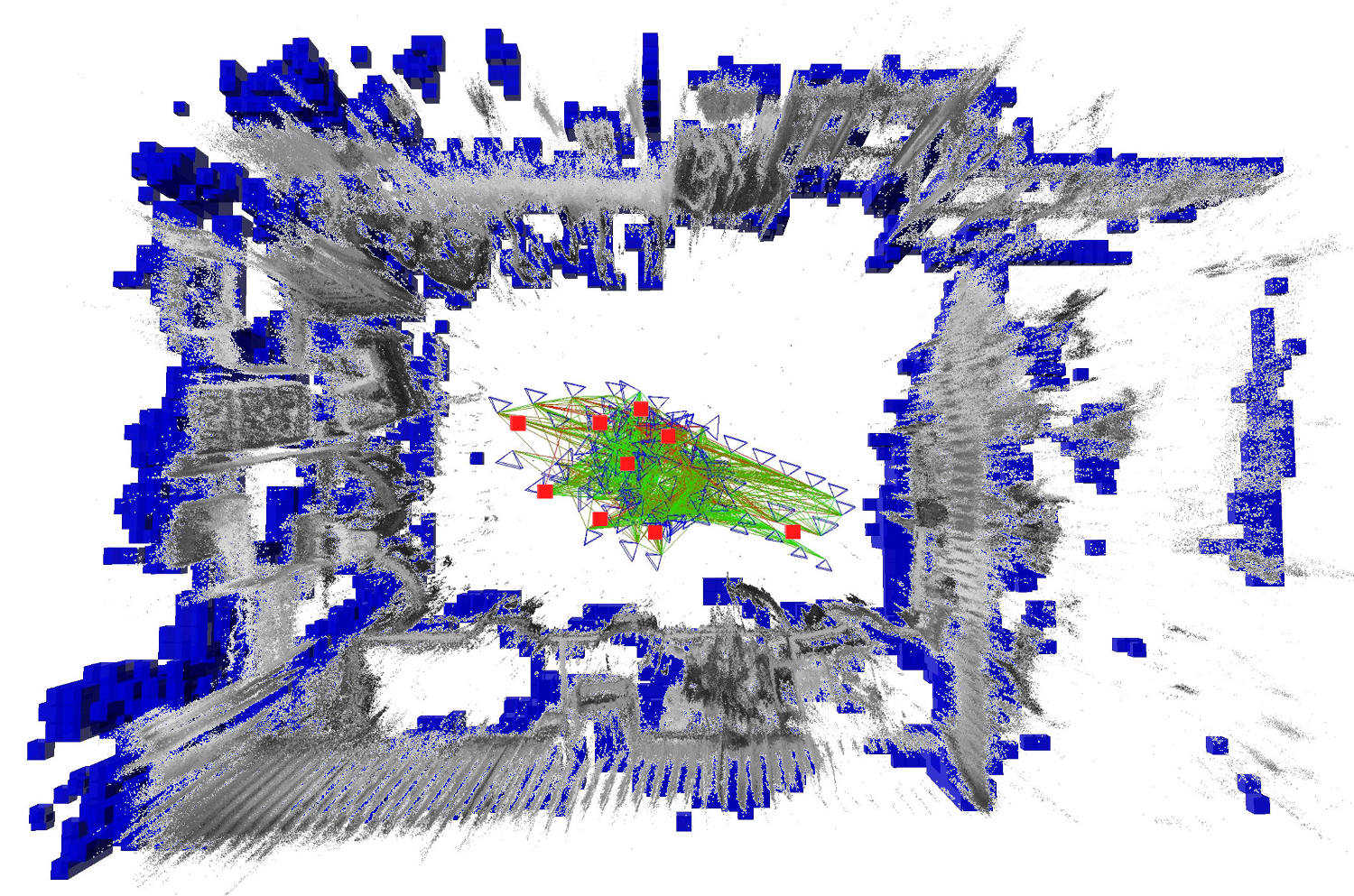}
        \caption{First experiment}
        \label{FigDiscoveryPlannedAndExecuted}
    \end{subfigure}\\
    %~ %add desired spacing between images, e. g. ~, \quad, \qquad, \hfill etc. 
    \begin{subfigure}[b]{0.89\linewidth}
        \includegraphics[width=\linewidth]{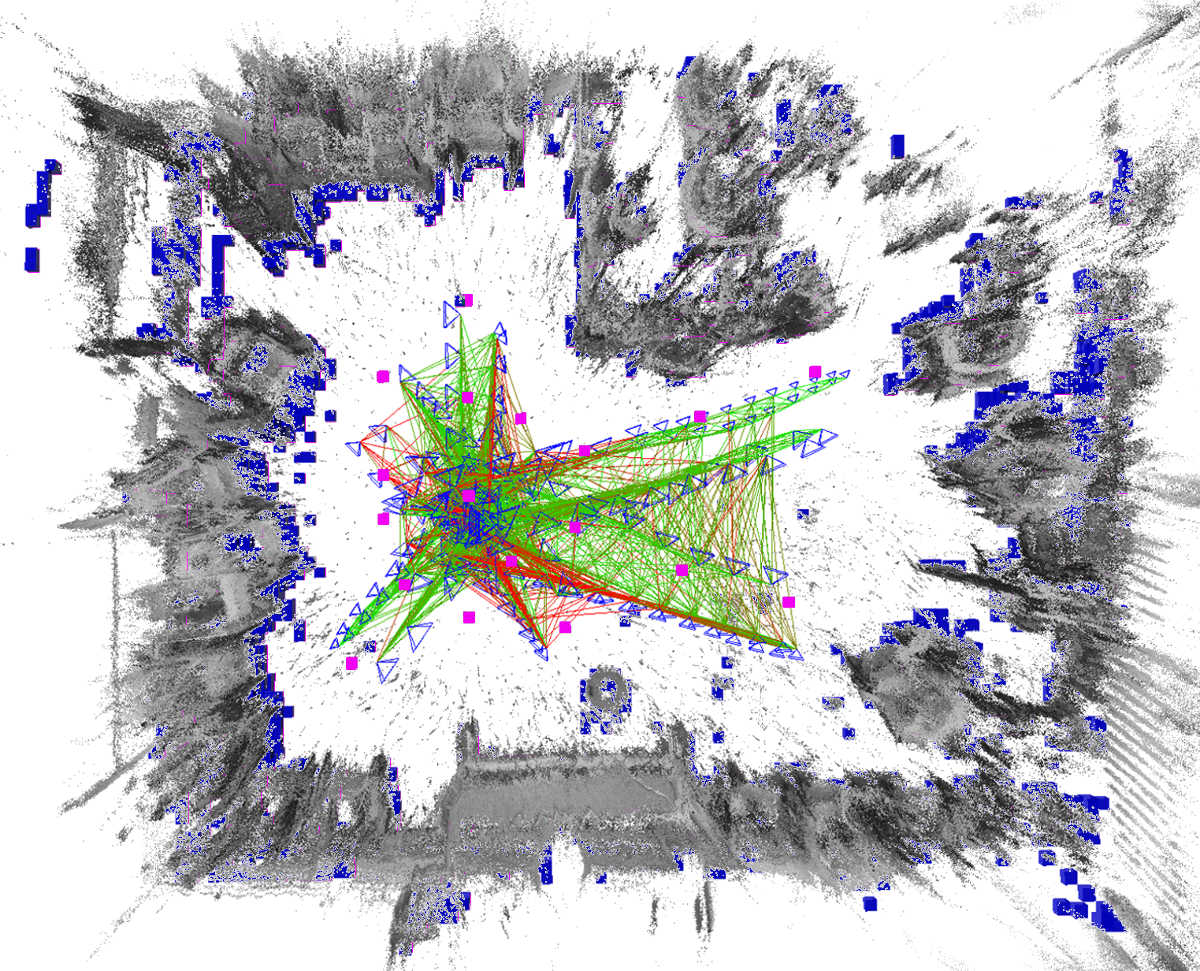}
        \caption{Second experiment}
        \label{FigEx3Discovery}
    \end{subfigure}
    \caption{3D occupancy map manually overlaid with semi-dense reconstruction, planned waypoints of the star discovery and executed trajectory (planned waypoints in red or pink, occupied voxels in blue, actual trajectory as estimated by LSD-SLAM as blue cameras, loop closure constraints as green lines).}
	\label{overlaidmaps}
\end{figure}

In the first experiment, we demonstrate autonomous local exploration using our star discovery strategy in our lab room.
There was no manual interaction except triggering the discovery and the landing at the end.
At first, the MAV performs a look-around maneuver.
In Fig.~\ref{FigLsdBeforeAfter} one can see the semi-dense reconstruction of the room obtained with LSD-SLAM.
Based on a 3D occupancy map, a star discovery is planned (Fig. \ref{FigDiscoveryPlan}).
In this case, we used three voxels in horizontal direction and one voxel in vertical direction to inflate the map. 

Fig.~\ref{FigDiscoveryPlannedAndExecuted} shows the planned waypoints of the star discovery overlaid with the actual trajectory estimate obtained with LSD-SLAM.
Fig.~\ref{FigOctomapBeforeAfter} shows how the executed trajectory has increased the number of free voxels in the occupancy map.

\subsubsection{Full Exploration Strategy}

\begin{figure}[tb]
    \centering
 \begin{subfigure}[t]{0.89\linewidth}
        \includegraphics[width=\linewidth]{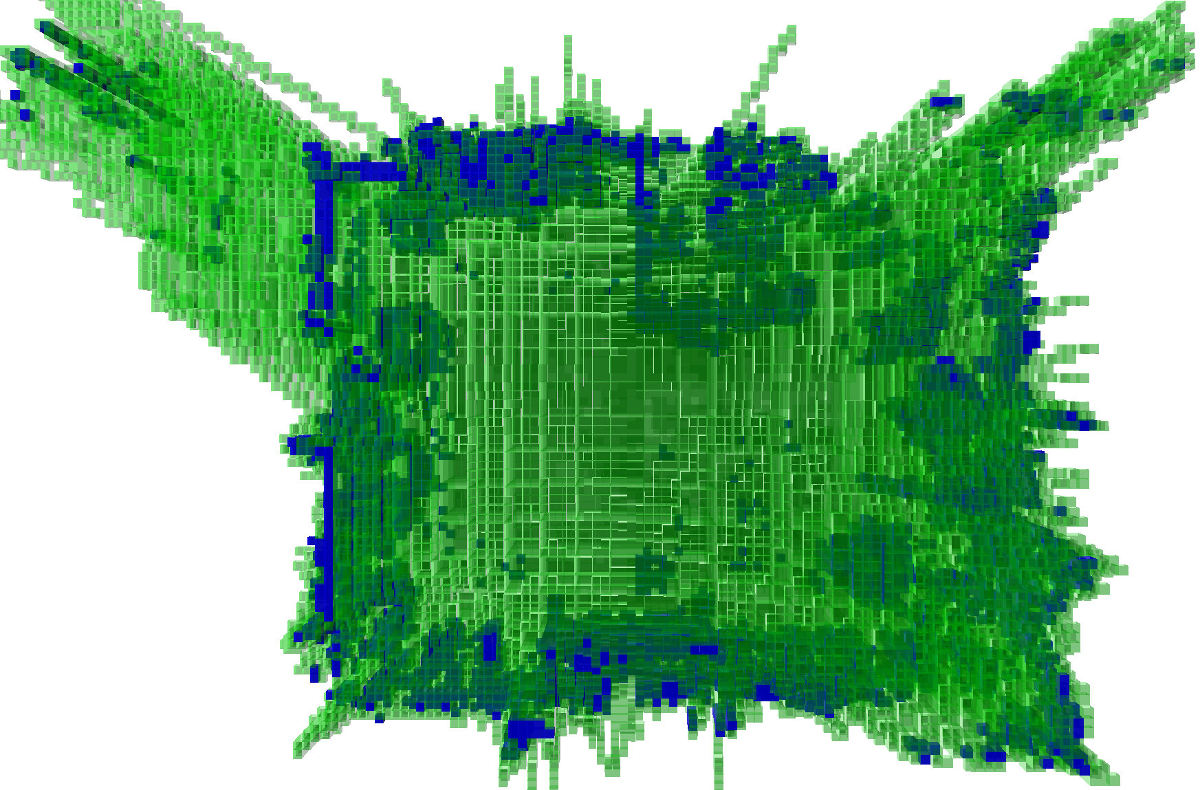}
        \caption{Free voxels (green)}
        \label{FigEx2Tree}
    \end{subfigure}\\
 \begin{subfigure}[t]{0.89\linewidth}
        \includegraphics[width=\linewidth]{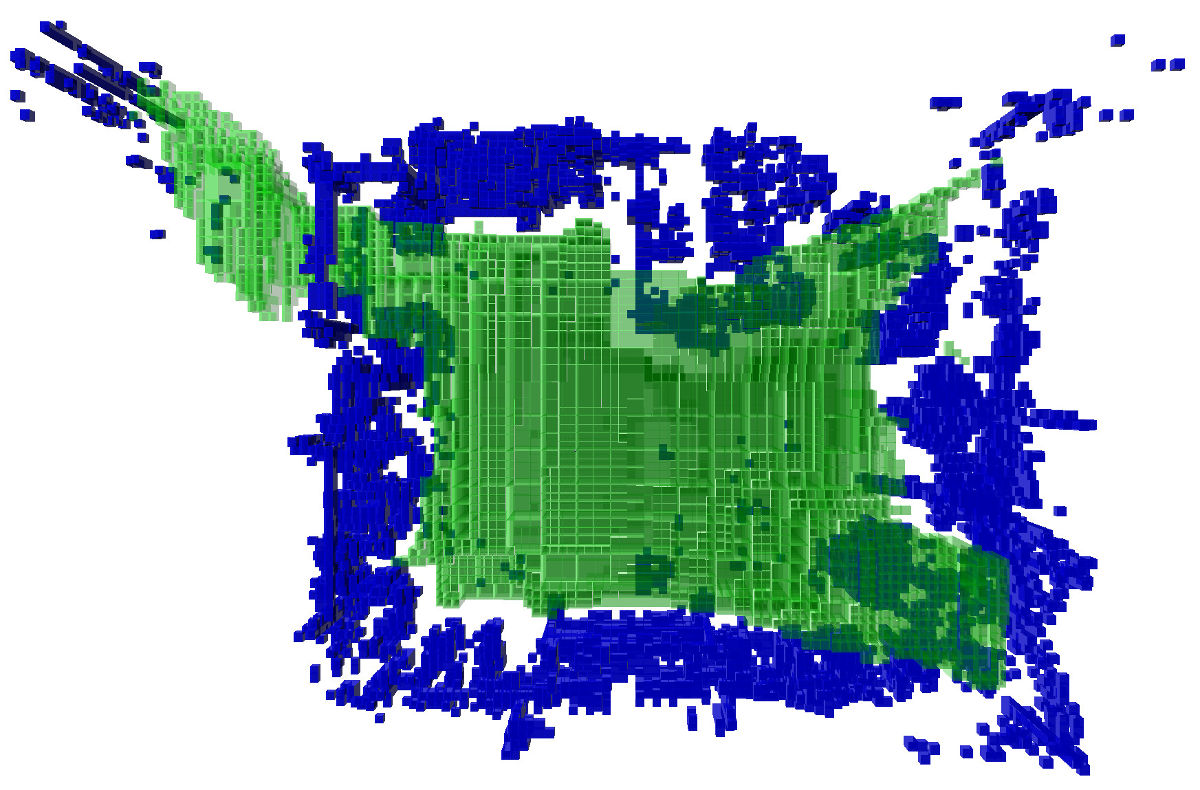}
        \caption{Traversable voxels (green) in the inflated map}
        \label{FigEx2ReallyFree}
    \end{subfigure}
    \caption{Occupied (blue), free and traversable voxels in the second experiment.}
	\label{FigEx2OctreesFirst}
\end{figure}

\begin{figure}[tb]
    \centering
    \begin{subfigure}[t]{0.89\linewidth}
        \includegraphics[width=\linewidth]{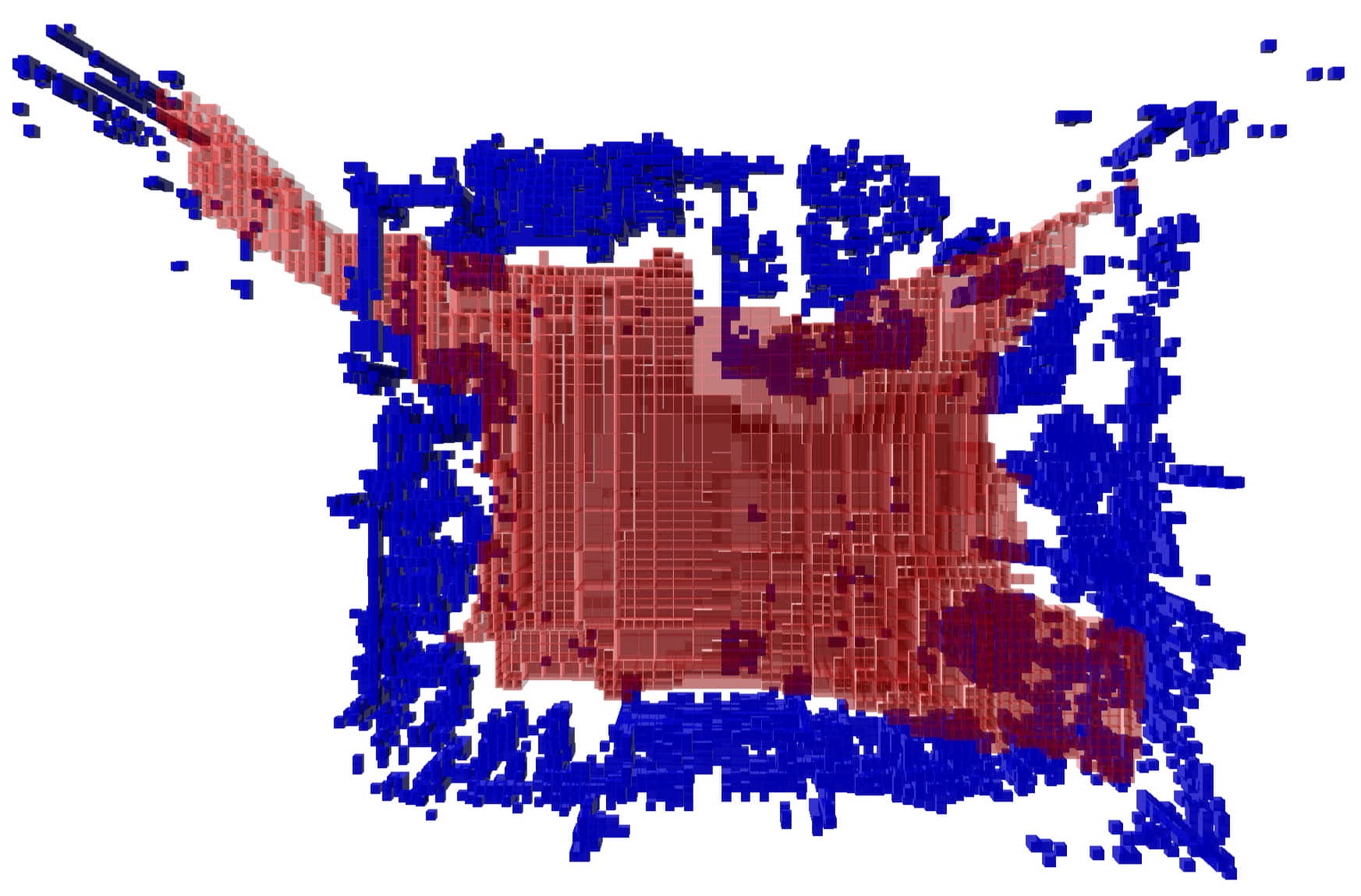}
        \caption{Traversable voxels (red) in line-of-sight from the star discovery origin.}
        \label{FigEx2ILOS}
    \end{subfigure}
    \begin{subfigure}[t]{0.89\linewidth}
        \includegraphics[width=\linewidth]{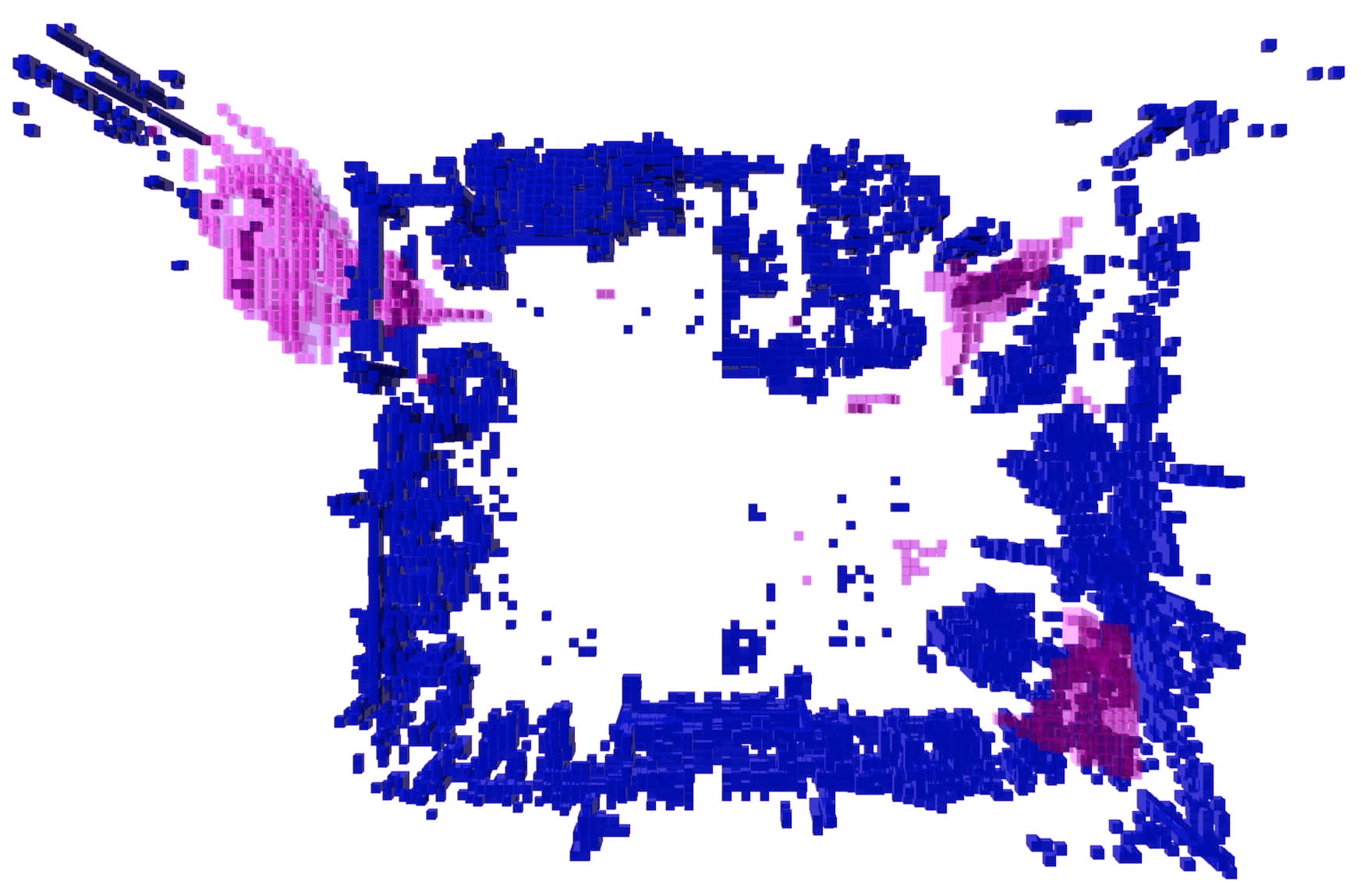}
        \caption{Interesting voxels (pink)}
        \label{FigEx2Interesting}
    \end{subfigure}
    \caption{Voxels in line of sight and interesting voxels in the second experiment.}
	\label{FigEx2OctreesSecond}
\end{figure}

\begin{figure}[htb]
    \centering
 \begin{subfigure}[t]{0.99\linewidth}
        \includegraphics[width=\linewidth]{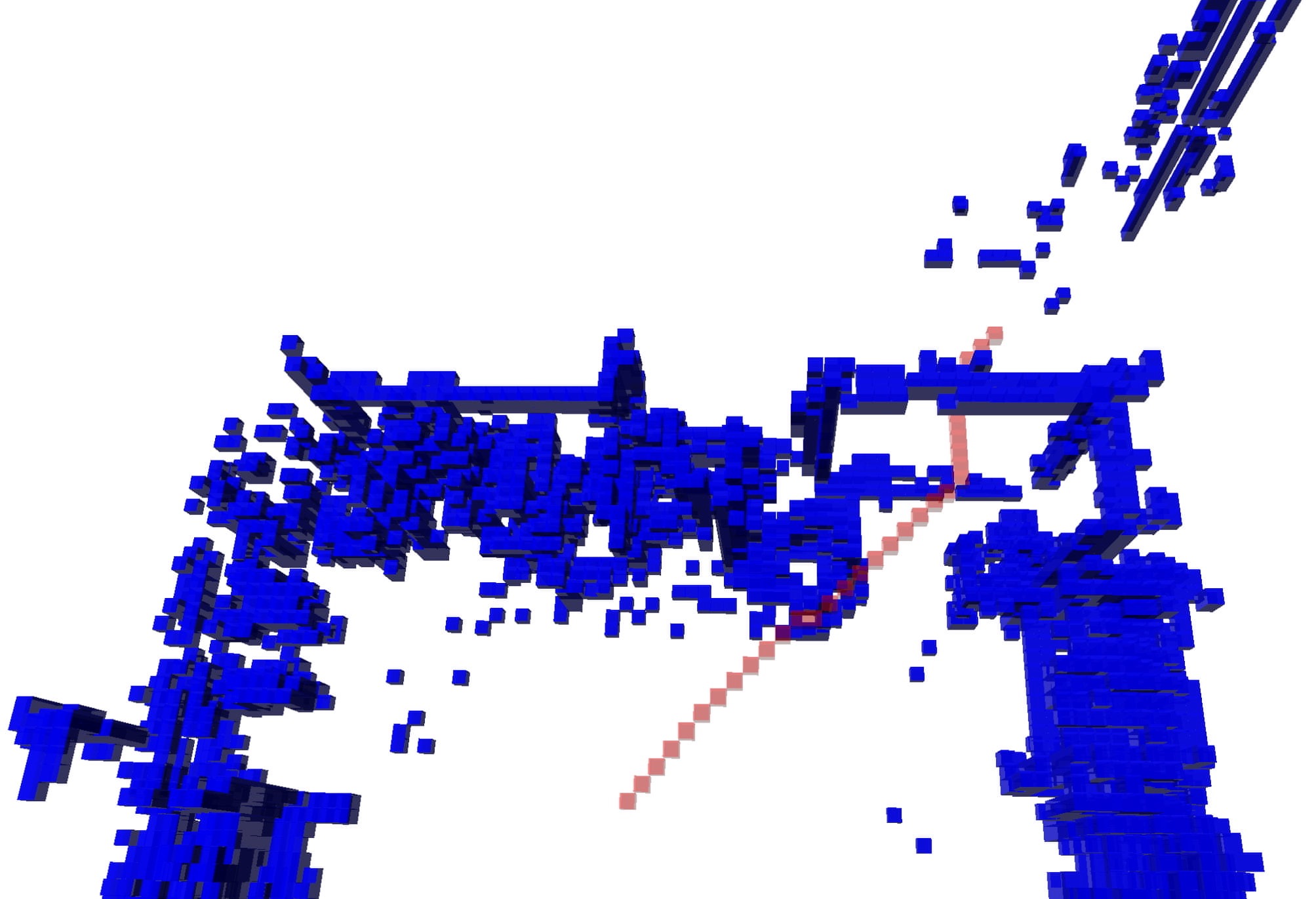}
        \caption{Planned path (red) to the found next star discovery origin.}
        \label{FigEx2Planned}
    \end{subfigure}\\
    %~ %add desired spacing between images, e. g. ~, \quad, \qquad, \hfill etc. 
    \begin{subfigure}[t]{0.99\linewidth}
        \includegraphics[width=\linewidth]{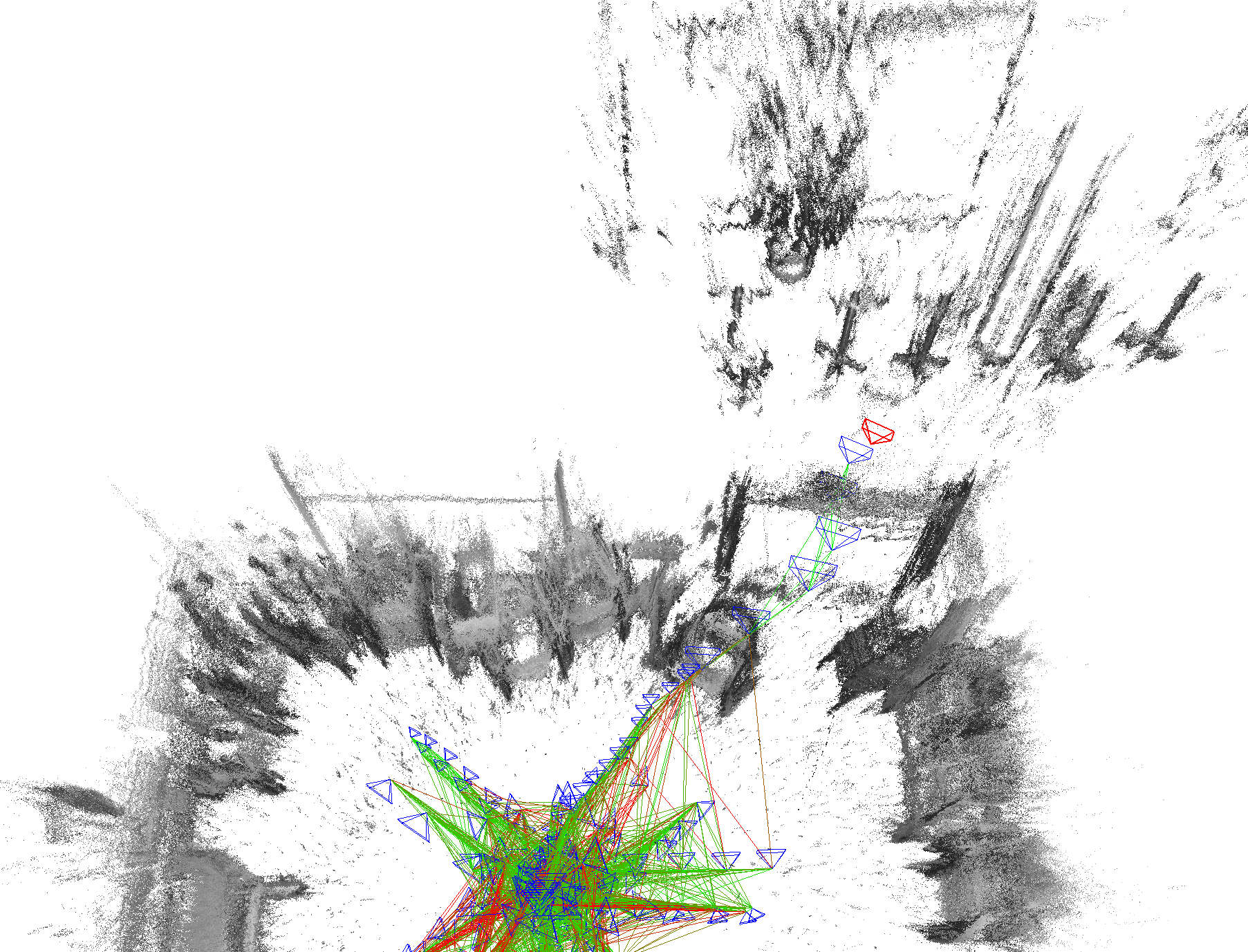}
        \caption{Flown trajectory (blue) to the interesting point estimated with LSD-SLAM.}
        \label{FigEx2Executed}
    \end{subfigure}
    %~ %add desired spacing between images, e. g. ~, \quad, \qquad, \hfill etc. 
    \caption{Planned and executed path in the second experiment.}
	\label{FigEx2PlannedAndExecuted}
\end{figure}

\begin{figure}[htb]      
  \begin{center}
    \includegraphics[width=0.85\linewidth]{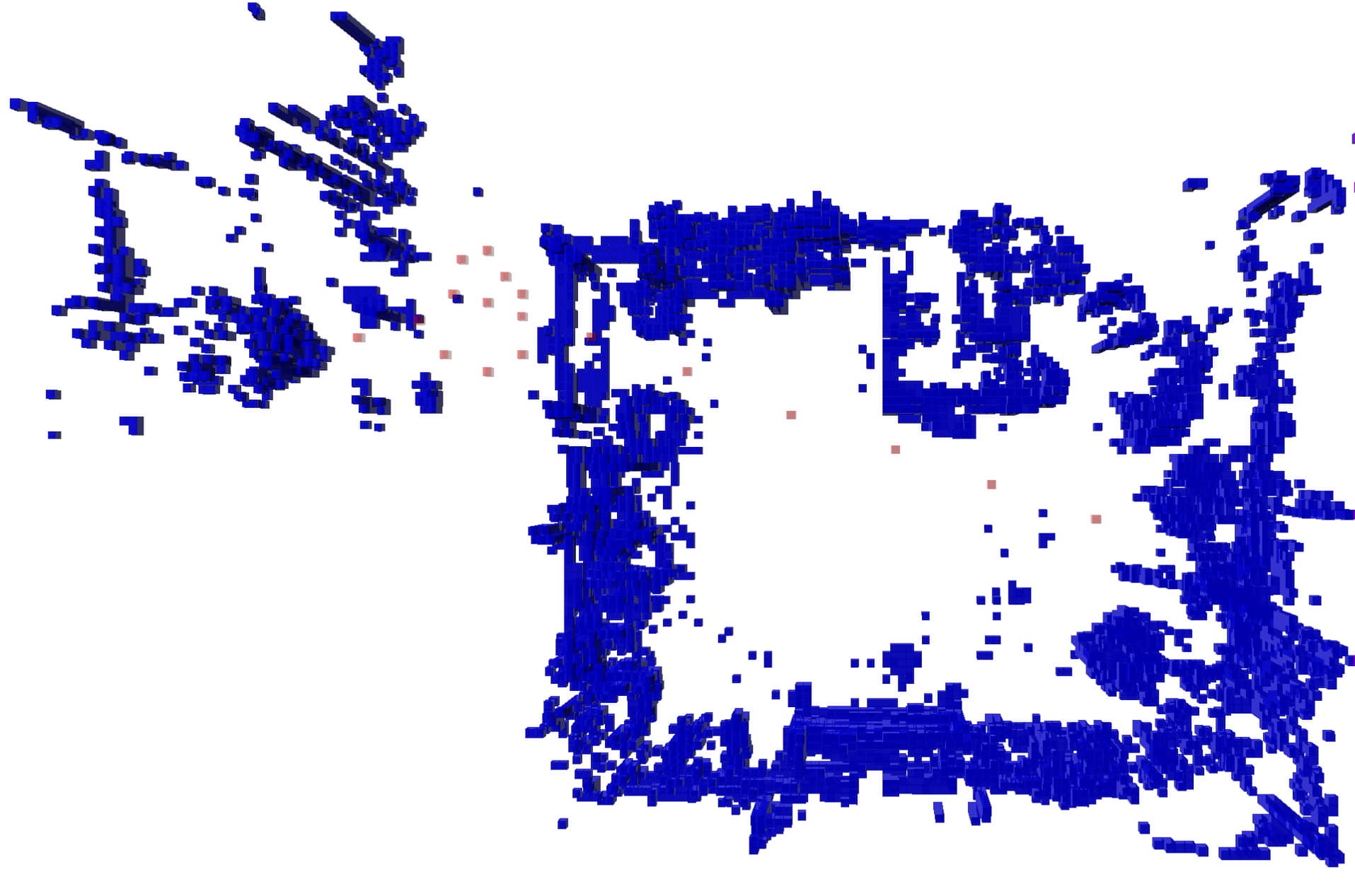}
  \end{center}
  \caption{Plan (red voxels) for the second star discovery in the second experiment.}
	\label{FigEx2Discovery2}
\end{figure}

\begin{table*}[ht]

\caption{Quantitative results on run-time and occupancy map statistics for the two experiments.}
\label{tab:quantitative_eval}
\begin{center}
  \begin{tabular}{c c c  c c c}
    \toprule
experiment & \multicolumn{2}{c}{1} & \multicolumn{3}{c}{2} \\
\cmidrule(rl){2-3} \cmidrule(rl){4-6}

    &look-around & star discovery & look-around & star discovery & new origin\\

occupancy map & 5.65s & 13.06s & 4.25s & 38.43s & 41.09s \\
inflating map & 0.69s & 0.76s & 1.40s & 2.15s & 2.86s \\
mark voxels in sight & - & - & - & 4.93s & 4.52/6.63s \\
way to new origin & - & - & - & 0.024s & 0.065s \\
\midrule
\#voxels in bounding box & 195048 & 211968 & 449565 & 728416 & 1312492 \\ 
\#free voxels & 36071 & 46021 & 75113 & 106944 & 159294 \\
\#occupied voxels & 8259 & 11477 & 6102 & 9673 & 10816 \\ 
\#free $\div$ \#known & 0.81 & 0.80 & 0.92 & 0.92 & 0.94 \\
\#free $\div$ \#bounding box & 0.18 & 0.22 & 0.17 & 0.15 & 0.12 \\ 
\#keyframes (approx.) & 66 & 162 & 54 & 236 & 257 \\
total \#points & 15411528 & 37828296 & 3152358 & 13776972 & 15002889 \\
    \bottomrule
  \end{tabular}
\end{center} 
\end{table*}

In the second experiment, we demonstrate a star discovery with subsequent repositioning at an interesting voxel in a larger seminar room.
First, the MAV took off, initialized the scale, and performed a look-around maneuver. 
Afterwards, the MAV executed a star discovery. 
Fig.~\ref{FigEx3Discovery} shows the planned discovery motion and the flown trajectory estimated with LSD-SLAM. 
We explain the differences by LSD-SLAM pose graph updates.

After the star discovery, we obtain the maps and interesting voxels in Fig.~\ref{FigEx2OctreesFirst} and Fig.~\ref{FigEx2OctreesSecond}. 
The largest connected component found by our algorithm is the one outside the room.
The MAV planned a path towards it and autonomously executed it.
In Fig.~\ref{FigEx2PlannedAndExecuted} we depicted the planned path and the actually flown trajectory estimated with LSD-SLAM.

After reaching the interesting point the battery of the MAV was empty and it landed automatically. The step that our algorithm would have performed next is the star discovery depicted in Fig.~\ref{FigEx2Discovery2}.

\subsection{Quantitative Evaluation}

Table~\ref{tab:quantitative_eval} gives results on the run-time of various parts of our approach and properties of the LSD-SLAM and occupancy mapping processes for the two experiments.
The creation of the occupancy map is visibly the most time-consuming part of our method, especially at later time steps when the semi-dense depth reconstruction becomes large.
In the second experiment modified parameters were used for the creation of the occupancy map. While they proved to perform better they also further increased the time consumption.
The remaining parts are comparatively time efficient and can be performed in a couple of seconds.
Our evaluation also shows that star discoveries significantly increase the number of free voxels in the map.

\section{CONCLUSIONS}

In this paper, we proposed a novel approach to vision-based navigation and exploration with MAVs.
Our method only requires a monocular camera, which enables low-cost, lightweight, and low-power consuming hardware solutions.
We track the motion of the camera and obtain a semi-dense reconstruction in real-time using LSD-SLAM.
Based on these estimates, we build 3D occupancy maps which we use for planning obstacle-free exploration maneuvers.

Our exploration strategy is a two-step process.
On a local scale, star discoveries find free-space in the local surrounding of a specific position in the map.
A global exploration strategy determines interesting voxels in the reachable free-space that is not in direct line-of-sight from previous star discovery origins.
In experiments, we demonstrate the performance of LSD-SLAM for vision-based navigation of a MAV.
We give qualitative insights and quantitative results on the effectiveness of our exploration strategy.

The success of our vision-based navigation and exploration method clearly depends on the robustness of the visual tracking.
If the MAV moves very fast into regions where it observes mostly textureless regions, tracking can become difficult.
A tight integration with IMU information could benefit tracking, however, such a method is not possible with the current wireless transmission protocoll for visual and IMU data on the Bebop. 

Also a more general path planning algorithm based on the next best view approach is desirable. This however requires a more efficient way to refresh the occupancy map when pose graph updates happen.

In future work we will extend our method to Stereo LSD-SLAM~\cite{engel15_stereo_lsdslam} and tight integration with IMUs.
We may also use the method for autonomous exploration on a larger MAV with onboard processing.

\addtolength{\textheight}{-2.8cm}   % This command serves to balance the column lengths on the last page

%%%%%%%%%%%%%%%%%%%%%%%%%%%%%%%%%%%%%%%%%%%%%%%%%%%%%%%%%%%%%%%%%%%%%%%%%%%%%%%%

\bibliographystyle{IEEEtran}

\bibliography{literature}

\begin{thebibliography}{10}
\providecommand{\url}[1]{#1}
\csname url@samestyle\endcsname
\providecommand{\newblock}{\relax}
\providecommand{\bibinfo}[2]{#2}
\providecommand{\BIBentrySTDinterwordspacing}{\spaceskip=0pt\relax}
\providecommand{\BIBentryALTinterwordstretchfactor}{4}
\providecommand{\BIBentryALTinterwordspacing}{\spaceskip=\fontdimen2\font plus
\BIBentryALTinterwordstretchfactor\fontdimen3\font minus
  \fontdimen4\font\relax}
\providecommand{\BIBforeignlanguage}[2]{{%
\expandafter\ifx\csname l@#1\endcsname\relax
\typeout{** WARNING: IEEEtran.bst: No hyphenation pattern has been}%
\typeout{** loaded for the language `#1'. Using the pattern for}%
\typeout{** the default language instead.}%
\else
\language=\csname l@#1\endcsname
\fi
#2}}
\providecommand{\BIBdecl}{\relax}
\BIBdecl

\bibitem{engel14eccv}
J.~Engel, T.~Sch\"ops, and D.~Cremers, ``{LSD-SLAM}: Large-scale direct
  monocular {SLAM},'' in \emph{ECCV}, 2014.

\bibitem{yamauchi1997_frontier}
B.~Yamauchi, ``A frontier-based approach for autonomous exploration,'' in
  \emph{IEEE Int. Symp. on Computational Intelligence in Robotics and
  Automation}, 1997.

\bibitem{gonzalesbanos2002_utility_exploration}
H.~H. Gonzalez-Banos and J.-C. Latombe, ``Navigation strategies for exploring
  indoor environments,'' \emph{Internaional Journal of Robotics Research},
  vol.~21, no. 10-11, pp. 829--848, 2002.

\bibitem{basilico2011_mcdm}
N.~Basilico and F.~Amigoni, ``Exploration strategies based on multi-criteria
  decision making for searching environments in rescue operations,''
  \emph{Autonomous Robots}, vol.~31, no.~4, pp. 401--417, 2011.

\bibitem{burgard2005_exploration}
W.~Burgard, M.~Moors, C.~Stachniss, and F.~Schneider, ``Coordinated multi-robot
  exploration,'' \emph{IEEE Transactions on Robotics}, vol.~21, no.~3, pp.
  376--386, 2005.

\bibitem{joho2007_3dexploration}
D.~Joho, C.~Stachniss, P.~Pfaff, and W.~Burgard, ``Autonomous exploration for
  3{D} map learning,'' in \emph{{A}utonome {M}obile {S}ysteme {(AMS)}}, 2007.

\bibitem{stachniss05robotics}
C.~Stachniss, G.~Grisetti, and W.~Burgard, ``Information gain-based exploration
  using rao-blackwellized particle filters,'' in \emph{Proc.~of Robotics:
  Science and Systems (RSS)}, Cambridge, MA, USA, 2005.

\bibitem{rekleitis2012_slurm}
I.~M. Rekleitis, ``Single robot exploration: Simultaneous localization and
  uncertainty reduction on maps (slurm),'' in \emph{CRV}, 2012.

\bibitem{shen2012_exploration}
S.~Shen, N.~Michael, and V.~Kumar, ``Autonomous indoor {3D} exploration with a
  micro-aerial vehicle,'' in \emph{IEEE International Conference on Robotics
  and Automation (ICRA)}, 2012, pp. 9--15.

\bibitem{yoder2015_surface_exploration}
L.~Yoder and S.~Scherer, ``Autonomous exploration for infrastructure modeling
  with a micro aerial vehicle,'' in \emph{Field and Service Robotics}, 2015.

\bibitem{nuske_jfr_2015}
S.~Nuske, S.~Choudhury, S.~Jain, A.~Chambers, L.~Yoder, S.~Scherer,
  L.~Chamberlain, H.~Cover, and S.~Singh, ``Autonomous exploration and motion
  planning for a {UAV} navigating rivers,'' \emph{Journal of Field Robotics},
  2015.

\bibitem{heng2015_mav_explore}
L.~Heng, A.~Gotovos, A.~Krause, and M.~Pollefeys, ``Efficient visual
  exploration and coverage with a micro aerial vehicle in unknown
  environments,'' in \emph{{IEEE} International Conference on Robotics and
  Automation (ICRA)}, 2015, pp. 1071--1078.

\bibitem{bircher2016_nbvplanner}
A.~Bircher, M.~Kamel, K.~Alexis, H.~Oleynikova, and R.~Siegwart, ``Receding
  horizon "next-best-view" planner for {3D} exploration,'' in \emph{IEEE
  International Conference on Robotics and Automation (ICRA)}, 2016, pp.
  1462--1468.

\bibitem{Desaraju2015}
V.~R. Desaraju, N.~Michael, M.~Humenberger, R.~Brockers, S.~Weiss, J.~Nash, and
  L.~Matthies, ``Vision-based landing site evaluation and informed optimal
  trajectory generation toward autonomous rooftop landing,'' \emph{Autonomous
  Robots}, vol.~39, no.~3, pp. 445--463, 2015.

\bibitem{engel14ras}
J.~Engel, J.~Sturm, and D.~Cremers, ``Scale-aware navigation of a low-cost
  quadrocopter with a monocular camera,'' \emph{Robotics and Autonomous Systems
  (RAS)}, vol.~62, no.~11, pp. 1646--–1656, 2014.

\bibitem{hornung2013_octomap}
A.~Hornung, K.~Wurm, M.~Bennewitz, C.~Stachniss, and W.~Burgard, ``{OctoMap}:
  an efficient probabilistic 3d mapping framework based on octrees,''
  \emph{Autonomous Robots}, vol.~34, no.~3, pp. 189--206, 2013.

\bibitem{engel15_stereo_lsdslam}
J.~Engel, J.~St\"uckler, and D.~Cremers, ``Large-scale direct {SLAM} with
  stereo cameras,'' in \emph{IROS}, 2015.

\end{thebibliography}

\end{document}